\documentclass[twoside,twocolumn,9pt]{article}
\usepackage{extsizes}
\usepackage[super,sort&compress,comma]{natbib} 
\usepackage[version=3]{mhchem}
\usepackage[left=1.5cm, right=1.5cm, top=1.785cm, bottom=2.0cm]{geometry}
\usepackage{balance}
\usepackage{mathptmx}
\usepackage{sectsty}
\usepackage{graphicx} 
\usepackage{lastpage}
\usepackage[format=plain,justification=justified,singlelinecheck=false,font={stretch=1.125,small,sf},labelfont=bf,labelsep=space]{caption}
\usepackage{float}
\usepackage{fancyhdr}
\usepackage{fnpos}
\usepackage[english]{babel}
\addto{\captionsenglish}{%
  
}
\usepackage{array}
\usepackage{droidsans}
\usepackage{charter}
\usepackage[T1]{fontenc}
\usepackage[usenames,dvipsnames]{xcolor}
\usepackage{setspace}
\usepackage[compact]{titlesec}
\usepackage{hyperref}

\usepackage{epstopdf}

\usepackage[section]{placeins}
\usepackage{multirow}

\definecolor{cream}{RGB}{222,217,201}

\begin{document}

\pagestyle{fancy}
\thispagestyle{plain}
\fancypagestyle{plain}{
\renewcommand{\headrulewidth}{0pt}
}

\makeFNbottom
\makeatletter
\renewcommand\LARGE{\@setfontsize\LARGE{15pt}{17}}
\renewcommand\Large{\@setfontsize\Large{12pt}{14}}
\renewcommand\large{\@setfontsize\large{10pt}{12}}
\renewcommand\footnotesize{\@setfontsize\footnotesize{7pt}{10}}
\makeatother

\renewcommand{\thefootnote}{\fnsymbol{footnote}}
\renewcommand\footnoterule{\vspace*{1pt}%
\color{cream}\hrule width 3.5in height 0.4pt \color{black}\vspace*{5pt}} 
\setcounter{secnumdepth}{5}

\makeatletter 
\renewcommand\@biblabel[1]{#1}            
\renewcommand\@makefntext[1]%
{\noindent\makebox[0pt][r]{\@thefnmark\,}#1}
\makeatother 
\renewcommand{\figurename}{\small{Fig.}~}
\sectionfont{\sffamily\Large}
\subsectionfont{\normalsize}
\subsubsectionfont{\bf}
\setstretch{1.125} 
\setlength{\skip\footins}{0.8cm}
\setlength{\footnotesep}{0.25cm}
\setlength{\jot}{10pt}
\titlespacing*{\section}{0pt}{4pt}{4pt}
\titlespacing*{\subsection}{0pt}{15pt}{1pt}

\fancyfoot{}
\fancyfoot[LO,RE]{\vspace{-7.1pt}\includegraphics[height=9pt]{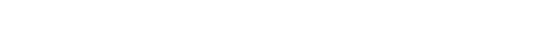}}
\fancyfoot[CO]{\vspace{-7.1pt}\hspace{13.2cm}\includegraphics{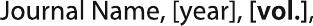}}
\fancyfoot[CE]{\vspace{-7.2pt}\hspace{-14.2cm}\includegraphics{head_foot/RF}}
\fancyfoot[RO]{\footnotesize{\sffamily{1--\pageref{LastPage} ~\textbar  \hspace{2pt}\thepage}}}
\fancyfoot[LE]{\footnotesize{\sffamily{\thepage~\textbar\hspace{3.45cm} 1--\pageref{LastPage}}}}
\fancyhead{}
\renewcommand{\headrulewidth}{0pt} 
\renewcommand{\footrulewidth}{0pt}
\setlength{\arrayrulewidth}{1pt}
\setlength{\columnsep}{6.5mm}
\setlength\bibsep{1pt}

\makeatletter 
\newlength{\figrulesep} 
\setlength{\figrulesep}{0.5\textfloatsep} 

\newcommand{\topfigrule}{\vspace*{-1pt}%
\noindent{\color{cream}\rule[-\figrulesep]{\columnwidth}{1.5pt}} }

\newcommand{\botfigrule}{\vspace*{-2pt}%
\noindent{\color{cream}\rule[\figrulesep]{\columnwidth}{1.5pt}} }

\newcommand{\dblfigrule}{\vspace*{-1pt}%
\noindent{\color{cream}\rule[-\figrulesep]{\textwidth}{1.5pt}} }

\makeatother

\twocolumn[
  \begin{@twocolumnfalse}
{\includegraphics[height=30pt]{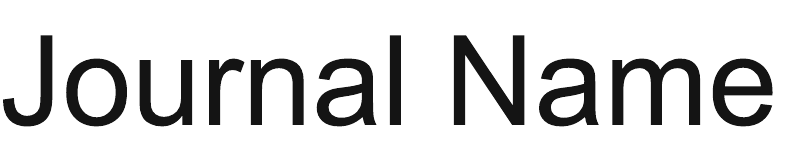}\hfill\raisebox{0pt}[0pt][0pt]{\includegraphics[height=55pt]{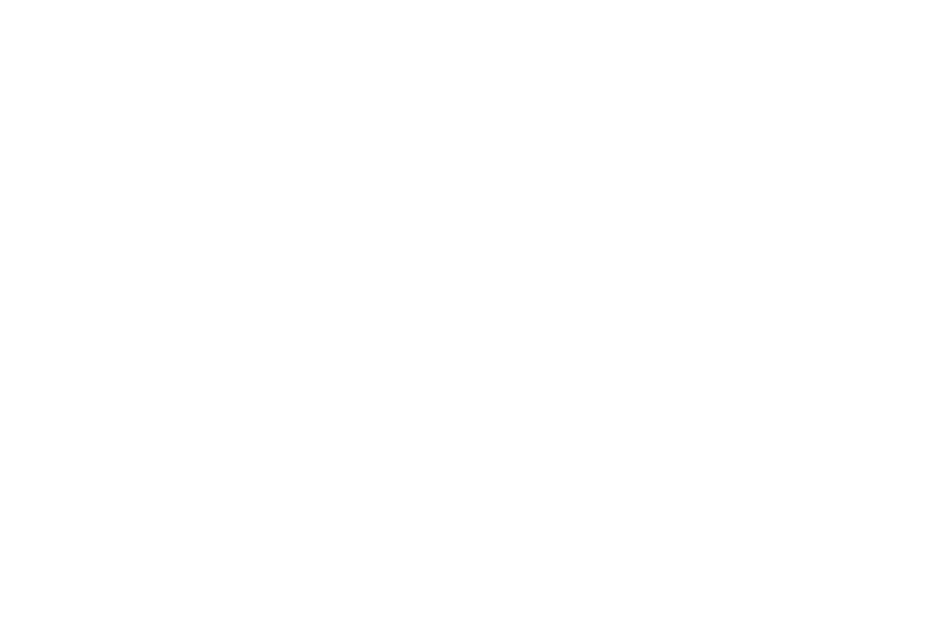}}\\[1ex]
\includegraphics[width=18.5cm]{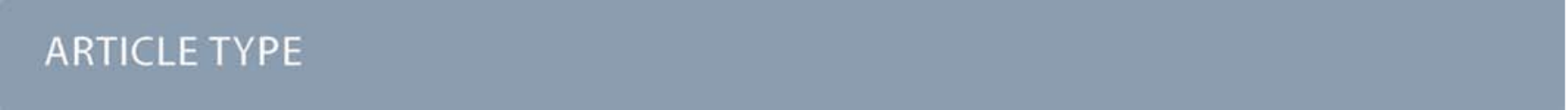}}\par
\vspace{1em}
\sffamily
\begin{tabular}{m{4.5cm} p{13.5cm} }

\includegraphics{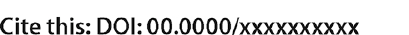} & \noindent\LARGE{\textbf{Can Deep Learning Assist Automatic Identification of Layered Pigments From XRF Data?}} \\
\vspace{0.3cm} & \vspace{0.3cm} \\

 & \noindent\large{Bingjie (Jenny) Xu\textit{$^{a}$}, Yunan Wu\textit{$^{a}$}, Pengxiao Hao\textit{$^{a, b}$}, Marc Vermeulen\textit{$^{a, c}$}, Alicia McGeachy\textit{$^{a, d}$}, Kate Smith\textit{$^{e}$}, Katherine Eremin\textit{$^{e}$}, Georgina Rayner\textit{$^{e}$}, Giovanni Verri\textit{$^{f}$}, Florian Willomitzer\textit{$^{a}$}, Matthias Alfeld\textit{$^{g}$}, Jack Tumblin\textit{$^{a}$}, Aggelos Katsaggelos\textit{$^{a}$} and Marc Walton\textit{$^{a, h\ast}$}} \\

\includegraphics{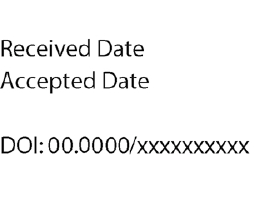} & \noindent\normalsize{X-ray fluorescence spectroscopy (XRF) plays an important role for elemental analysis in a wide range of scientific fields, especially in cultural heritage. XRF imaging, which uses a raster scan to acquire spectra across artworks, provides the opportunity for spatial analysis of pigment distributions based on their elemental composition. However, conventional XRF-based pigment identification relies on time-consuming elemental mapping by expert interpretations of measured spectra. To reduce the reliance on manual work, recent studies have applied machine learning techniques to cluster similar XRF spectra in data analysis and to identify the most likely pigments. Nevertheless, it is still challenging for automatic pigment identification strategies to directly tackle the complex structure of real paintings, e.g. pigment mixtures and layered pigments. In addition, pixel-wise pigment identification based on XRF imaging remains an obstacle due to the high noise level compared with averaged spectra.
Therefore, we developed a deep-learning-based end-to-end pigment identification framework to fully automate the pigment identification process. In particular, it offers high sensitivity to the underlying pigments and to the pigments with a low concentration, therefore enabling satisfying results in mapping the pigments based on single-pixel XRF spectrum. As case studies, we applied our framework to lab-prepared mock-up paintings and two 19th-century paintings: Paul Gauguin's \textit{Poèmes Barbares} (1896) that contains layered pigments with an underlying painting, and Paul Cezanne's \textit{The Bathers} (1899-1904). The pigment identification results demonstrated that our model achieved comparable results to the analysis by elemental mapping, suggesting the generalizability and stability of our model. } \\

\end{tabular}

 \end{@twocolumnfalse} \vspace{0.6cm}
]

\renewcommand*\rmdefault{bch}\normalfont\upshape
\rmfamily
\section*{}
\vspace{-1cm}

\footnotetext{\textit{$^{a}$~Northwestern University, 2145 Sheridan Road, Evanston, IL, United States.}}

\footnotetext{\textit{$^{b}$~Shanghai University, Shanghai, 200444, China.}}

\footnotetext{\textit{$^{l}$~The National Archives, Bessant Dr, Richmond TW9 4DU, United Kingdom.}}

\footnotetext{\textit{$^{d}$~The Metropolitan Museum of Art, 1000 5th Ave, New York, NY, United States.}}

\footnotetext{\textit{$^{e}$~Harvard Art Museums, Straus Center for Conservation and Technical Studies, 32 Quincy St, Cambridge, MA, United States.}}

\footnotetext{\textit{$^{f}$~Art Institute of Chicago, 111 S Michigan Ave, Chicago, IL, United States.}}

\footnotetext{\textit{$^{g}$~Delft University of Technology, 2628 CN Delft, Netherlands.}}
\footnotetext{\textit{$^{h}$~M+, 38 Museum Drive, West Kowloon Cultural District, Hong Kong, China.}}
\footnotetext{\textit{$^{\ast}$~Corresponding author: marc.walton@mplus.org.hk}}






\section{Introduction}
X-ray fluorescence spectroscopy (XRF) is a well-established workhorse technique for elemental analysis in a wide range of scientific fields~\cite{Vanhoof2021}, such as geochemistry~\cite{rowe2012quantification, oyedotun2018x, sarala2016comparison}, forensic science~\cite{langstraat2017large, nakano2011depth} and archaeology~\cite{shackley2011introduction}. Few areas of research benefit from its use as much as the investigation of cultural heritage that often necessitates \textit{in-situ} investigations under ambient conditions as is offered by open-architecture and hand-held versions of the XRF instrument. To fully characterize the heterogeneous nature and complex history of many artworks, it often requires a high number of measurements that are best implemented in the form of XRF imaging. To form an XRF image, the instrument moves across the surface of an object, such as a painting, while collecting spectra point-by-point that are spatially redressed to their 2-D locations~\cite{Alfeld2017}. The XRF imaging allows for spatial analysis of pigment distributions based on their elemental composition.

Conventional XRF-based pigment identification uses underlying spectrum evaluation methods~\cite{van1993spectrum} to generate elemental maps, which existing XRF analysis software, for example, PyMCA~\cite{sole2007multiplatform}, can fully support. However, identifying the pigments that cause these elemental maps and their spectra requires experts with prior knowledge of the painting, as most artworks consist of a considerable variety of pigments in varying pigment mixtures or arranged in complicated stratigraphy~\cite{alfeld2015strategies, romano2017real, alfeld2013revealing}. To assist and reduce these experts' manual work, machine learning techniques have recently been applied to simplify pigment identification by clustering pigment-related spectral features~\cite{kogou2021new}. For example, XRFast, an open-source unsupervised sparse dictionary learning algorithm developed recently, finds maps of correlated elements to help in pigment identification, an improvement on the traditional approach that seeks correlations by image processing~\cite{XRFast}. 
Today, deep learning (DL) has been widely applied to assist with XRF analysis and has the potential to perform fully automatic identification of elements and their sources. For example, Shugar et al. identified 48 different wood species with the XRF dataset using convolutional neural networks, reaching an accuracy of $99\%$~\cite{shugar2021rapid}. They found 0.7-1.7 keV the most important portion of the spectra for wood classification, which involves elements of calcium, aluminum and magnesium. Moreover, Kim et al. applied a neural network on micro XRF data to generate mineral maps on natural rocks. They showed that DL was a good way to improve the description of mineral reactivity to rock samples of different origin, size, and thickness~\cite{kim2022quantification}. Most recently, Cerys et al. proposed a deep-learning-based method to directly identify pigments from XRF spectra~\cite{jones2022neural}. By training a convolutional neural network, they classified XRF spectra into one of the 15 pigment classes with an high accuracy, but claimed that it was still challenging to apply the model to more complex scenarios, including layered pigments and pigment mixtures. Therefore, focusing on the complicated stratigraphies of real paintings, this work built convolutional neural networks to automatically identify pigment mixtures in layered structures and to display 2D pigment maps.

We propose an end-to-end pigment identification framework, including pigment library creation, XRF spectra simulation, mock-up preparation, a pigment identification DL model, and 2D pigment map generation. As a case study, we applied our framework to a 19th-century painting, Paul Gauguin's \textit{Poèmes Barbares} (1896), focusing on a set of 19-century pigments previously identified in this painting~\cite{Vermeulen2021}. In addition, previous analysis revealed a hidden painting beneath the surface, which tremendously increased the difficulty of pigment identification.
Therefore, our DL model targeted on the multilayered structure of the painting. By training the DL model using 16224 simulated XRF spectra of three-layered pigments, followed by finetuning the model using $20\%$ of the experimental XRF spectra (i.e., 1320 XRF spectra) of mock-up paintings, the DL model demonstrated satisfying performance of pigment identification on the mock-ups as well as \textit{Poèmes Barbares}, particularly showing a high sensitivity of underlying pigments or the pigments with a low concentration.

To further demonstrate the applicability of this approach, we applied the finetuned model to Paul Cezanne's \textit{The Bathers} (1899-1904). This is a single-layered painting created from a comparable time period with \textit{Poèmes Barbares}, composed of similar but fewer types of pigments. The DL model achieved satisfying ability to identify the pigments of \textit{The Bathers}, suggesting our model's generalizability and stability.

In all, our framework provides an automatic and quick pigment identification strategy based on non-invasive XRF imaging, in particular targeting the paintings' complex layered structure. Given the trained model, it does not require expertise or extensive familiarity working with XRF and pigments, but directly answers where the pigment might exist. Although the type of pigments is still limited in the current work, our framework shows great potential for extension to other types of pigments and paintings, as well as XRF-based identification problems in the fields beyond cultural heritage.

\section{Experimental Method}
We gathered XRF data from three sources to build, train, and test our framework: XRF spectra from existing paintings, XRF spectra from our own oil-paint 'mock-ups' of crossed paint stripes of different pigment mixtures, and simulated XRF spectra of multilayered pigments.

\subsection{XRF Datasets}
Here, we selected Paul Gaugin's \textit{Poèmes Barbares} (1896) to test our pigment identification approach. To some extent, this painting could represent the challenging pigment identification task of many 19th-century paintings, which often involve a wide range of pigments made available by the industrial revolution. Furthermore, the types of pigments and the structure of the paint layers of \textit{Poèmes Barbares} were studied previously by combining XRF, reflectance imaging spectroscopy and cross-section analysis~\cite{Vermeulen2021}, therefore providing us with reliable ground-truth measurements of this historical work. 

These studies show that \textit{Poèmes Barbares} consists of a visible (top) painting and a hidden (bottom) painting, each constructed with multiple paint layers and various mixtures of pigments\cite{Vermeulen2021}. To accommodate the complex layer structure for the DL approach, we simplified it into a three-layer structure: one top pigment layer, one bottom pigment layer, and one ground layer. Referring to the chemical analysis of multiple cross-sections of the painting, we chose 11 distinct pigments of interest (Table \ref{tbl:pigment}) and calcium carbonate as the ground layer.  

Next, we made our own set of layered oil paintings to better measure how the layers affect the pigments' XRF spectra. For these layered-painting 'mock-ups' (Figure \ref{fig:Structure}), we chose a reasonable range of the pigment fractions, binder ratios, and layer thicknesses found in \textit{Poèmes Barbares}. We then created sets of 3-layer oil paintings consisting of crossed strips of paints with these values and prepared them for XRF measurements. In addition, we generated a simulation dataset for these same pigments and thicknesses to train and validate the deep learning model.

\subsubsection{Experimental Dataset}
\begin{figure}[htp!]
 \centering
 \includegraphics[height=5cm]{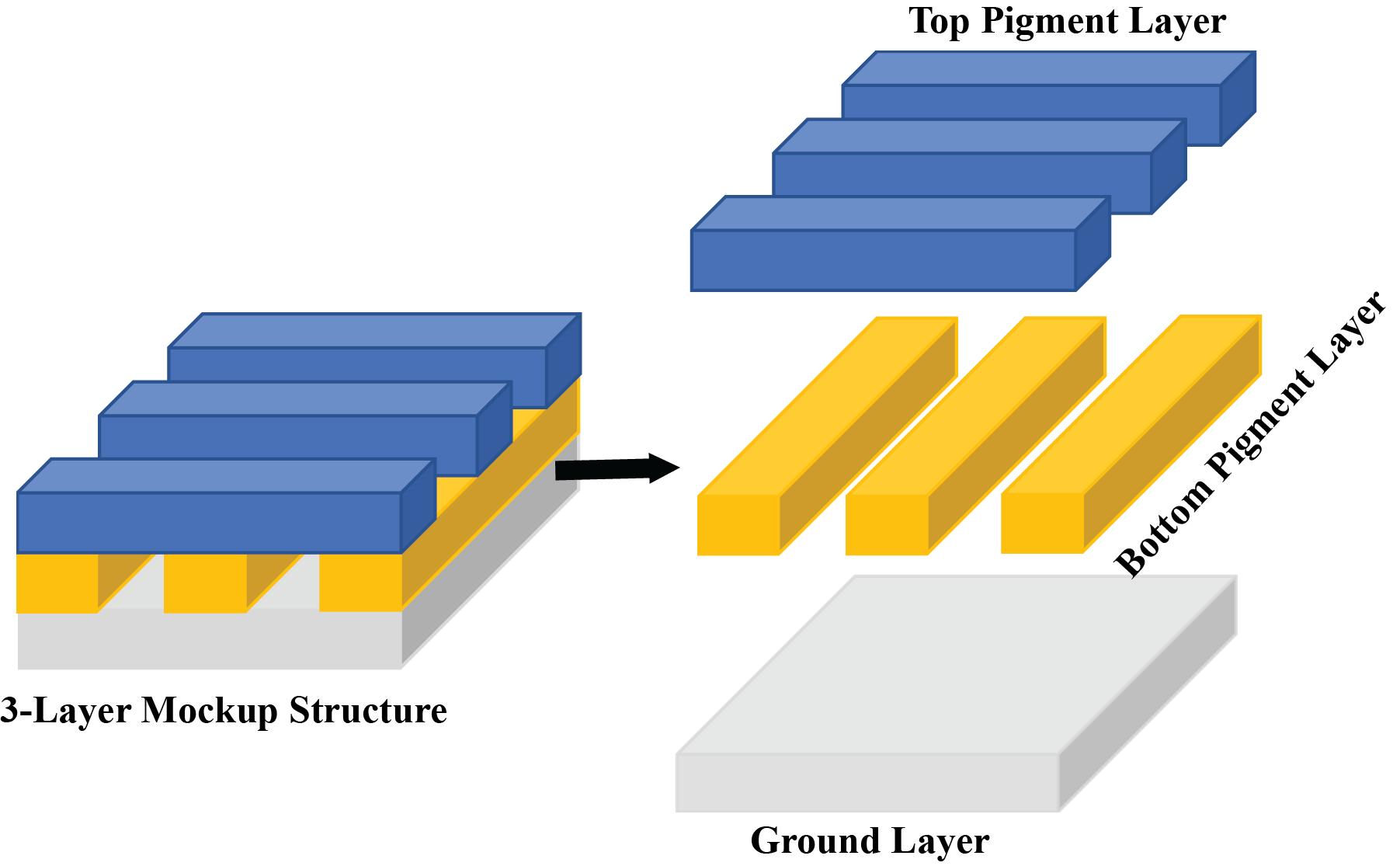}
 \caption{The structure of mock-up samples. The first and second layers both contained multiple strips of pigment layers, each strip with its pigment combination selected from Table \ref{tbl:pigment}; the third layer was a ground layer consisting of calcium carbonate.} 
 \label{fig:Structure}
\end{figure}

The experimental dataset contained XRF spectra of mock-ups with known pigment layer structure as ground truth and was used for training and testing the deep learning model. We prepared the three-layered mock-ups with various combinations of pigments: six pigment mixtures as the bottom paint layer (mainly varying in the pigments' mass fractions, providing 16 bottom layers in total) and four mixtures as the top paint layer, making up 64 different layer structures (Table \ref{tbl:mockup}). Fig. \ref{fig:Structure} illustrated the structure of our mock-up samples: the first and second layer both contained multiple strips of pigment layers, where the combination of pigments in each strip was selected from Table \ref{tbl:mockup}; the third layer was a fixed ground layer of calcium carbonate.

A tape casting coater (model MSK-AFA-HC100, MTI Corporation (Richmond, CA)) was used to deposit the paint layer sequentially with adjustable thicknesses. A new layer was painted after the previous layer completely dried. In all mock-ups, we applied a calcium carbonate ground layer of 150 -- 200 $\mu$m thickness to match that of \textit{Poèmes Barbares}. As for the pigment layers, the type of mixtures simulated the palette of both the bottom and top painting in \textit{Poèmes Barbares}. To include the XRF effects commonly found in layered systems and pigment mixtures, such as shielding and matrix effects, we varied the pigment fractions in the bottom paint layers (e.g. Bottom 1A, Bottom 1B, Bottom 1C in Table \ref{tbl:mockup}), while the top paint layer differed in its layer thickness (30 -- 200 $\mu$m). 

In preparing the mock-ups, commercial pigments and the binder were hand-ground for 10 min to obtain a uniform mixture\cite{Sturdy2016}. Since the organic binder would not significantly affect the XRF signal, we used Galkyd Lite (Gamblin Artists Colors (Portland, OR)) for its fast-drying property. A pigment-to-binder ratio (w\%:w\%) of 3:1 was applied to achieve the mobility required by the tape casting coater. For lead-containing paint mixtures which appeared dryer, we added drops of Gamsol odorless mineral spirits (Gamblin Artists Colors) to further dilute the mixture. All pigment mixtures were deposited on pH-neutral art boards (Crescent (Wheeling, IL)). 

To collect the experimental XRF dataset, we scanned the mock-ups with the XGLab ELIO XRF imaging spectrometer system. The XRF spectra of the mock-ups were acquired at 40 kV and 40 $\mu$A. We set the acquisition time at 10.0 sec per point to increase the signal-to-noise ratio required for deep learning. The raster scan was executed using a 100 x 100 mm motorized X-Y linear stage mount (Zaber T-LSM100A) with a step size of 1 x 1 mm.

\begin{table}[htp!]
\small
  \caption{\ Pigment library for selected pigments}
  \label{tbl:pigment}
  \begin{tabular*}{0.48\textwidth}{@{\extracolsep{\fill}}lll}
    \hline
    Index & Pigment\textsuperscript{a} & Chemical Formula \\
    \hline
    1 & Calcium carbonate & CaCO$_3$ \\
    2 & Chrome oxide green (CrG) & {Cr}$_2\textnormal{O}_3$ \\
    3 & Chrome yellow (CrY) & PbCrO$_4$ \\
    4 & Cobalt blue (CB) & CoO\textnormal{\textperiodcentered}\textnormal{Al}$_2\textnormal{O}_3$\\
    5 & Emerald green (EG)\textsuperscript{b}  & Cu(CH$_3\textnormal{COO)}_2 \textnormal{\textperiodcentered}  \textnormal{3Cu(AsO}_2\textnormal{)}_2$\\
    6 & Iron oxide (IO) & Fe$_2\textnormal{O}_3$ \\
    7 & Lead white (LW) & 2PbCO$_3 \textnormal{\textperiodcentered}\textnormal{Pb(OH)}_2$ \\
    8 & Prussian blue (PB) & Fe[Fe$_2\textnormal{(CN)}_6$]$_3$ \\
    9 & Red lead (RL) & Pb$_3\textnormal{O}_4$ \\
    10 & Carmine (CM)\textsuperscript{c} & SnO$_2$ \\
    11 & Vermilion (VM) & HgS \\
    12 & Zinc white (ZW) & ZnO \\
    \hline
  \end{tabular*}
  \emph{a} Calcium carbonate, chrome oxide green, cobalt blue, verdigris, iron oxide red (120 M), lead white, Prussian blue, and vermilion were purchased from Kremer Pigmente (New York, NY). Chrome yellow and red lead were purchased from Rublev Colours (Willits, CA). Zinc oxide was obtained from Gamblin Artists Colors (Portland, OR). Sodium arsenite ($\mathrm{\geq 90\%}$) and tin oxide ($\mathrm{99.99\%}$)  were obtained from Sigma Aldrich. 
  
  \emph{b} Due to the current unavailability of commercial emerald green pigment, we mixed the copper carbonate pigment verdigris (VG, $\mathrm{Cu(CH}_3$$\mathrm{COO)}_2$\textperiodcentered$\mathrm{2Cu(OH)}_2$) and sodium arsenite (SA, $\mathrm{NaAsO}_2$) to approximate the XRF signal of emerald green, in which the Cu-As mass ratio was set accordingly.
  
  \emph{c} Since the chemical analysis suggested that tin oxide ($\mathrm{SnO}_2$) was the support of carmine, we solely used the $\mathrm{SnO}_2$ powder in preparing the mock-ups to represent carmine in the XRF dataset.
\end{table}

\begin{table}[htp!]
\small
  \caption{\ Summary of pigment mixtures in the mock-ups}
  \label{tbl:mockup}
  \begin{tabular*}{0.48\textwidth}{@{\extracolsep{\fill}}lll}
    \hline
    Pigment Mixture & Compound\textsuperscript{a} & Mass fraction (\%) \\
    \hline
    Top 1 & LW, CM, PB, CB & 55, 10, 25, 10 \\
    Top 2 & VM, CM, CB & 30, 30, 40 \\
    Top 3 & VM, CrY, IO, VG, SA & 39, 6, 39, 6, 10 \\
    Top 4 & VM & 100 \\
    Bottom 1A & ZW, LW, RL, VM & 10, 10, 10, 70 \\
    Bottom 1B & ZW, LW, RL, VM & 10, 20, 50, 20 \\
    Bottom 1C & ZW, LW, RL, VM & 10, 50, 20, 20 \\
    Bottom 2A & ZW, LW, VM & 10, 85, 5 \\
    Bottom 2B & ZW, LW, VM & 10.7, 88.7, 0.6 \\
    Bottom 2C & ZW, LW, VM & 10, 50, 40 \\
    Bottom 3A & ZW, LW, CB, CrG & 10, 15, 15, 60 \\
    Bottom 3B & ZW, LW, CB, CrG & 10, 15, 60, 15 \\
    Bottom 3C & ZW, LW, CB, CrG & 10, 60, 15, 15 \\
    Bottom 4A & ZW, LW, CrY, VG, SA & 10, 45, 15, 12, 18 \\
    Bottom 4B & ZW, LW, CrY, VG, SA & 10, 30, 30, 12, 18 \\
    Bottom 4C & ZW, LW, CrY, VG, SA & 10, 15, 15, 24, 36 \\
    Bottom 5A & ZW, VM, CrY, VG, SA & 10, 5, 5, 32, 48 \\
    Bottom 5B & ZW, VM, CrY, VG, SA & 78, 6, 6, 4, 6 \\
    Bottom 5C & ZW, VM, CrY, VG, SA & 10, 15, 15, 24, 36 \\
    Bottom 6 & VM & 100 \\
    \hline
  \end{tabular*}
  
  \emph{a} The pigments and their corresponding abbreviations used in this manuscript are listed as below:
  CB: cobalt blue, CrG: chrome oxide green, CM: carmine, CrY: chrome yellow, EG: emerald green, IO: iron oxide, LW: lead white, PB: Prussian blue, RL: red lead, SA: sodium arsenite, VG:verdigris, VM: vermilion, ZW: zinc white.
  
\end{table}

\subsubsection{Simulation Dataset}
To ensure a sufficient dataset size for training our DL model, we generated a simulation dataset of 16224 XRF spectra in total. The spectra were calculated using the \texttt{matrixSpectrum} function in \texttt{PyMca5.PyMcaGui.physics.xrf.} \texttt{McaAdvancedFit} of the PyMCA Python package~\cite{sole2007multiplatform}. Based on the fundamental parameter approach~\cite{de2009multilayers}, the function simulates XRF spectra for multilayer samples using a multilayer quantitative XRF analysis algorithm. In generating the simulation dataset, we applied a three-layer structure (top pigment layer - bottom pigment layer - ground layer) similar with the mock-up paintings. Each of the top and the bottom layers consisted of a single pigment from the pigment library (Table \ref{tbl:pigment}), with a layer thickness of 50 -- 200 $\mu$m (10 $\mu$m interval) and 100 -- 150 $\mu$m (10 $\mu$m interval), respectively. 

\subsection{Data Preprocessing} \label{data_preprocessing}

Before feeding the experimental and simulation datasets into the DL model, several preprocessing steps were necessary. First, based on preliminary ablation studies, the overlaps between the sulfur-K lines (2.31 keV) and the lead-M (2.34 keV) or mercury-M lines (2.20 keV) confused the DL model in distinguishing the three elements. As a result, both the experimental and simulation datasets started from 2.80 keV to improve the performance of the DL model. 

Also, the simulated XRF spectra lacked the underlying spectral background signal caused by X-ray scattering and the equipment properties. Therefore, we estimated this spectral background from the mean of the measured spectra using SNIP~\cite{RYAN1988396} and added it to the simulation dataset to better mimic their experimental counterpart, without significantly changing the key elemental peaks.

Since the XRF sensitivity to different elements varies, both the experimental and simulated spectra exhibited skewness of more than one order of magnitude, while the elemental concentrations were generally comparable. To reduce this skewness, a log-log square root transformation (Eq.\ref{eq:eq1}) of the original spectrum $X$ followed by normalization to $X'$ was applied. This normalization step further enhanced the recognition of the elements with low peak intensities in the spectrum. $X'$ was the final input features to the DL model.

\begin{equation}
X' = \log_e(1 + {\log_e(1 + {\sqrt{X}})}),
\label{eq:eq1}
\end{equation}

\subsection{Model Architecture}

\begin{figure*}[htp!] 
\begin{center}
\begin{tabular}{l} 
\includegraphics[width=\textwidth]{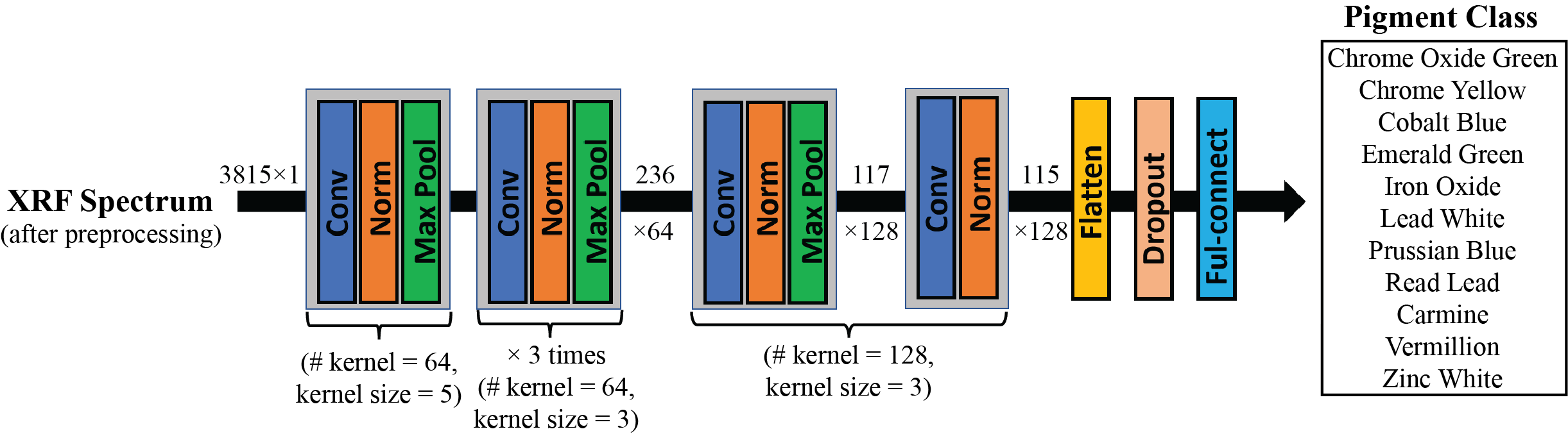}
\end{tabular}
\end{center}
\caption{The Architecture of the Deep Learning Model. It contained 1D convolutional layers (Conv), normalization layers (Norm), max pooling layers (Max Pool), a flatten layer, a dropout layer and a fully-connected layer. The input size of each XRF spectrum was 3815$\times$1. The output predictions were in 11 classes (one per pigment). \# kernel and kernel size stood for the number and the size of the kernel at the corresponding convolutional layer, respectively. }
{\label{fig:model} }
\end{figure*}

The model consisted of 5 convolutional blocks, where each block was made up of a 1D convolutional layer, an activation function LeakyReLU~\cite{xu2015empirical}, a batch normalization layer, and a max-pooling layer, as shown in Fig. \ref{fig:model}. According to preliminary ablation studies, the number and size of the kernels of each 1D convolutional layer were set at 64,64,64,64,128 and 5,3,3,3,3, respectively. The activation function LeakyReLU, i.e., $f(x) = max(0.01x, x)$, kept the positive part of its input while preventing "the dead ReLU issue" by using a small value when the input was negative~\cite{xu2015empirical}. The batch normalization layer sped up the training process by distributing the input for every layer around the same mean and standard deviation. The max-pooling layer downsampled the dimension of the input to half. Finally, the model was followed by a post-convolutional layer with 128 kernels with a size of 3, a normalization layer, a dropout layer with a rate of 0.25, and one fully-connected layer of 11 classes, outputting the probabilities of each class (each pigment) between 0 to 1. The output layer used a sigmoid activation function $\sigma(z_j)= \frac{1}{1 + e^{-z_j}}$, where $z_j$ was the predicted score from the model of each class. The probabilities as predicted further built the pigment maps of the paintings. Two datasets, the simulation and experimental datasets, were used to train the model and compared. The two training processes applied the same architecture as described but differed in the initial weights. The model trained with the simulation dataset used randomly initialized weights. It was then applied as a pre-trained model, its weights used as the initial value and further finetuned in training the model with the experimental dataset, a strategy known as Transfer Learning (TL)\cite{zhuang2020comprehensive}.

TL is a popular technique that uses the pre-trained weights from an initial model as the starting point on another model, which reduces or eliminates the risk of overfitting and allows for better training speed and model performance~\cite{raghu2019transfusion}. Therefore in this paper, we pre-trained the model on the simulation dataset and then refined it using the the experimental dataset. Specifically, when training with the experimental dataset, the pre-trained weights from the simulation dataset were first fixed (i.e., untrainable) in all convolutional layers, whereas only the fully connected layers were finetuned with the targeted dataset. Next, all layers were trainable and were further finetuned with the experimental dataset. The model performances with and without finetuning were compared in the result section to show the effectiveness of TL.            

Moreover, the loss was calculated to optimize the performance of the model by averaging the binary cross entropy of each predicted class, as defined in Eq.\ref{eq:eq2}.
\begin{equation}
Loss = -\frac{1}{N}\sum_{i=1}^{N}\sum_{j=1}^{K}[\hat{z}_j\cdot log(z_j) + (1-\hat{z}_j)\cdot log(1-z_j)] ,
\label{eq:eq2}
\end{equation}
where $\hat{z}$ was the ground truth label, $z$ was the score predicted from the model for each class, $K$ was the number of the class, and $N$ was the number of batch size.

\subsection{Training Strategy}
The training process was completed in two steps. The model proposed in Fig. \ref{fig:model} was first pre-trained on the simulation dataset. A total of 16224 simulated XRF spectra were splitted into a testing dataset and a training dataset with a ratio of 1:4. In training the model, we applied the five-fold cross-validation \cite{stone1974cross} by further dividing the training dataset into '$5$' groups of equal size and iteratively selecting one group as the validation set, while the rest remaining as the training set. Therefore, with the full iteration, the model's performance was evaluated by the testing dataset five times. All cohorts of the dataset were preprocessed following section \ref{data_preprocessing}. The model was trained with the Adam optimizer~\cite{kingma2014adam} with an initial learning rate of 0.001. Adam was chosen for the model due to its robustness, less convergence time and fewer parameters for tuning. The batch size was 64, and on average, it took about 0.95 hours for each fold in the 5-fold cross-validation for 150 epochs with early stopping settings. Next, the experimental dataset had 6604 XRF spectra from the two-layer pigment areas, which we manually picked from all mock-ups with known ground truth. To confirm the effect of finetuning on pigment identification, we tested the model before and after finetuning by the experimental dataset. Specifically, to test the model before finetuning, which was trained with the simulation dataset, all experimental datasets were used as the testing dataset. The model after fintuning was initialized with the weights that performed the best among the five-fold cross-validation and was further finetuned with $20\%$ of the randomly selected data from the experimental dataset, i.e. 1320 XRF spectra. The remaining $80\%$ (5284 XRF spectra) was used as the testing dataset. Similar to the training process of the model trained with the simulation dataset, the finetuned model was trained with the Adam optimizer with a lower learning rate of 0.0005. The batch size was 64, and it took an average of 0.2 hours for each of the 5 groups or 'folds'. All training and testing processes were performed on an NVIDIA GeForce RTX 2070 GPU using Tensorflow 2.0 in Python 3.7. 

\section{Results}
\subsection{Pigment identification model performance}

\begin{table*}[ht]
\small
  \caption{\ The classification results for each pigment class among the models trained from three different datasets: the simulation dataset, the experimental dataset without finetuning, and the experimental dataset acquired from the mock-ups with finetuning\textsuperscript{a}}
  \label{tbl:result}
  \begin{tabular*}{\textwidth}{@{\extracolsep{\fill}}llllllllll}
    \hline
    \multirow{2}{*}{\textbf{Pigment Class}}         & \multicolumn{3}{c}{\textbf{Simulation}}                                 & \multicolumn{3}{c}{\textbf{Experimental (no finetune)}} & \multicolumn{3}{c}{\textbf{Experimental (finetune)}} \\ 
    & \multicolumn{1}{l}{Accuracy} & \multicolumn{1}{l}{Sensitivity} & F1    & \multicolumn{1}{l}{Accuracy}          & \multicolumn{1}{l}{Sensitivity}          & F1            & \multicolumn{1}{l}{Accuracy}         & \multicolumn{1}{l}{Sensitivity}         & F1           \\ \hline
    Cobalt Blue            & \multicolumn{1}{l}{0.950}    & \multicolumn{1} {l}{0.997}       & 0.782 & \multicolumn{1} {l}{0.870}             & \multicolumn{1} {l}{0.973}                & 0.890         & \multicolumn{1} {l}{0.899}            & \multicolumn{1} {l}{0.985}               & 0.916        \\
    Emerald Green          & \multicolumn{1} {l}{0.964}    & \multicolumn{1} {l}{1.0}         & 0.861 & \multicolumn{1} {l}{0.773}             & \multicolumn{1} {l}{0.821}                & 0.820         & \multicolumn{1} {l}{0.859}            & \multicolumn{1} {l}{0.870}              & 0.871        \\
    Iron Oxide             & \multicolumn{1} {l}{0.882}    & \multicolumn{1} {l}{0.732}       & 0.687 & \multicolumn{1} {l}{0.659}             & \multicolumn{1} {l}{0.506}                & 0.550         & \multicolumn{1} {l}{0.998}            & \multicolumn{1} {l}{0.995}               & 0.997        \\
    Prussian Blue          & \multicolumn{1} {l}{0.878}    & \multicolumn{1} {l}{0.572}       & 0.592 & \multicolumn{1} {l}{0.869}             & \multicolumn{1} {l}{0.771}                & 0.746         & \multicolumn{1} {l}{0.996}            & \multicolumn{1} {l}{0.994}               & 0.993        \\
    Carmine                   & \multicolumn{1} {l}{1.0}      & \multicolumn{1} {l}{1.0}         & 1.0   & \multicolumn{1} {l}{0.653}             & \multicolumn{1} {l}{0.664}                & 0.685         & \multicolumn{1} {l}{0.995}            & \multicolumn{1} {l}{0.992}               & 0.995        \\
    Vermilion             & \multicolumn{1} {l}{0.987}    & \multicolumn{1} {l}{0.997}       & 0.955 & \multicolumn{1} {l}{0.918}             & \multicolumn{1} {l}{0.994}                & 0.952         & \multicolumn{1} {l}{0.947}            & \multicolumn{1} {l}{0.978}               & 0.970        \\
    Zinc White                    & \multicolumn{1} {l}{0.952}    & \multicolumn{1} {l}{1.0}         & 0.813 & \multicolumn{1} {l}{0.916}             & \multicolumn{1} {l}{0.918}                & 0.956         & \multicolumn{1} {l}{0.916}            & \multicolumn{1} {l}{0.953}               & 0.954        \\
    Chrome Yellow        & \multicolumn{1} {l}{0.933}    & \multicolumn{1} {l}{0.981}       & 0.708 & \multicolumn{1} {l}{0.631}             & \multicolumn{1} {l}{0.642}                & 0.687         & \multicolumn{1} {l}{0.831}            & \multicolumn{1} {l}{0.862}               & 0.842        \\
    Chrome Oxide Green         & \multicolumn{1} {l}{0.942}    & \multicolumn{1} {l}{0.982}       & 0.781 & \multicolumn{1} {l}{0.563}             & \multicolumn{1} {l}{0.569}                & 0.587         & \multicolumn{1} {l}{0.798}            & \multicolumn{1} {l}{0.887}               & 0.620        \\
    Red Lead               & \multicolumn{1} {l}{0.871}    & \multicolumn{1} {l}{0.653}       & 0.626 & \multicolumn{1} {l}{0.508}             & \multicolumn{1} {l}{0.458}                & 0.556         & \multicolumn{1} {l}{0.868}            & \multicolumn{1} {l}{0.687}               & 0.634        \\
    Lead White             & \multicolumn{1} {l}{0.858}    & \multicolumn{1} {l}{0.634}       & 0.629 & \multicolumn{1} {l}{0.755}             & \multicolumn{1} {l}{0.767}                & 0.860         & \multicolumn{1} {l}{0.836}            & \multicolumn{1} {l}{0.957}               & 0.886        \\ \hline
    \end{tabular*}
     \emph{a} The results sum up the number of pigment class predictions of both the top- and bottom-layered pigments. The results are averaged from five-fold cross-validation.
\end{table*}

Table \ref{tbl:result} shows the effectiveness of different training approaches for pigment classification; first trained solely with simulation data, then with experimental data without finetuning, then with the experimental data with finetuning. The overall accuracy, sensitivity, and F1 score were calculated for each class (each pigment) averaged from the five fold validation groups of experimental data on the testing datasets. Accuracy and Sensitivity are defined in Eq.\ref{eq:acc} and \ref{eq:sen}.
\begin{equation}
Accuracy = \frac{TP + TN}{TP + TN + FN +FP},
\label{eq:acc}
\end{equation}

\begin{equation}
Sensitivity = \frac{TP}{TP + FN}.
\label{eq:sen}
\end{equation}
F1 score (the harmonic mean of precision and sensitivity) evaluates the imbalanced classes, as defined in Eq.\ref{eq:f1}.
\begin{equation}
F1 = \frac{2 * Precision * Sensitivity}{Precision + Sensitivity},
\label{eq:f1}
\end{equation}
where 
\begin{equation}
Precision = \frac{TP}{TP + FP}.
\label{eq:pre}
\end{equation}
The output predictions are classified into TP, TN, FP, and FN, which are short for True Positive, True Negative, False Positive, and False Negative, respectively. True or False denotes whether the class exists or not according to the ground truth. Positive or negative suggests if the (pigment) class is predicted as existing or non-existing. All four evaluation metrics range from 0 to 1, and the closer to 1, the better the performance of the model. The analyses are performed on the same GPU using the scikit-learn package in Python 3.7.     

As shown in Table \ref{tbl:result}, the model trained from the simulation dataset generally provided satisfactory accuracy, which ranges from 0.858 to 1.0. Sensitivity varies  from 0.572 to 1.0, and F1 score varies from 0.592 to 1.0. According to the sensitivity and F1 score, two groups of pigments - iron-containing Prussian blue and iron oxide, and lead-containing red lead and lead white - perform worse than other pigments due to the similar elemental profiles within each group. This model can be generalized to the experimental data without finetuning but with relatively worse performance on carmine (tin-based, according the the cross-section analysis), chrome oxide green, and chrome yellow. Compared with the simulation data, tin has a much lower concentration in carmine~\cite{Dapson2007}, causing a bigger error and therefore a lower accuracy for the experimental data. In the simulation dataset, there is no pigment mixture. Chrome oxide green only contains Cr and chrome yellow contains Cr and Pb, which helps the model to separate them. But Pb also exists in other pigments in the mock-ups, which confuses the model. However, finetuning the model on the experimental data significantly improves the classification results, reaching an overall accuracy ranging from 0.798 to 0.998, sensitivity from 0.687 to 0.995, and F1 from 0.634 to 0.997. The next section will further explain our finetuning strategy.

\subsection{Tests on Paul Gauguin's \textit{Poèmes Barbares}} 
\begin{figure}[htp!]
 \centering
 \includegraphics[height=5cm]{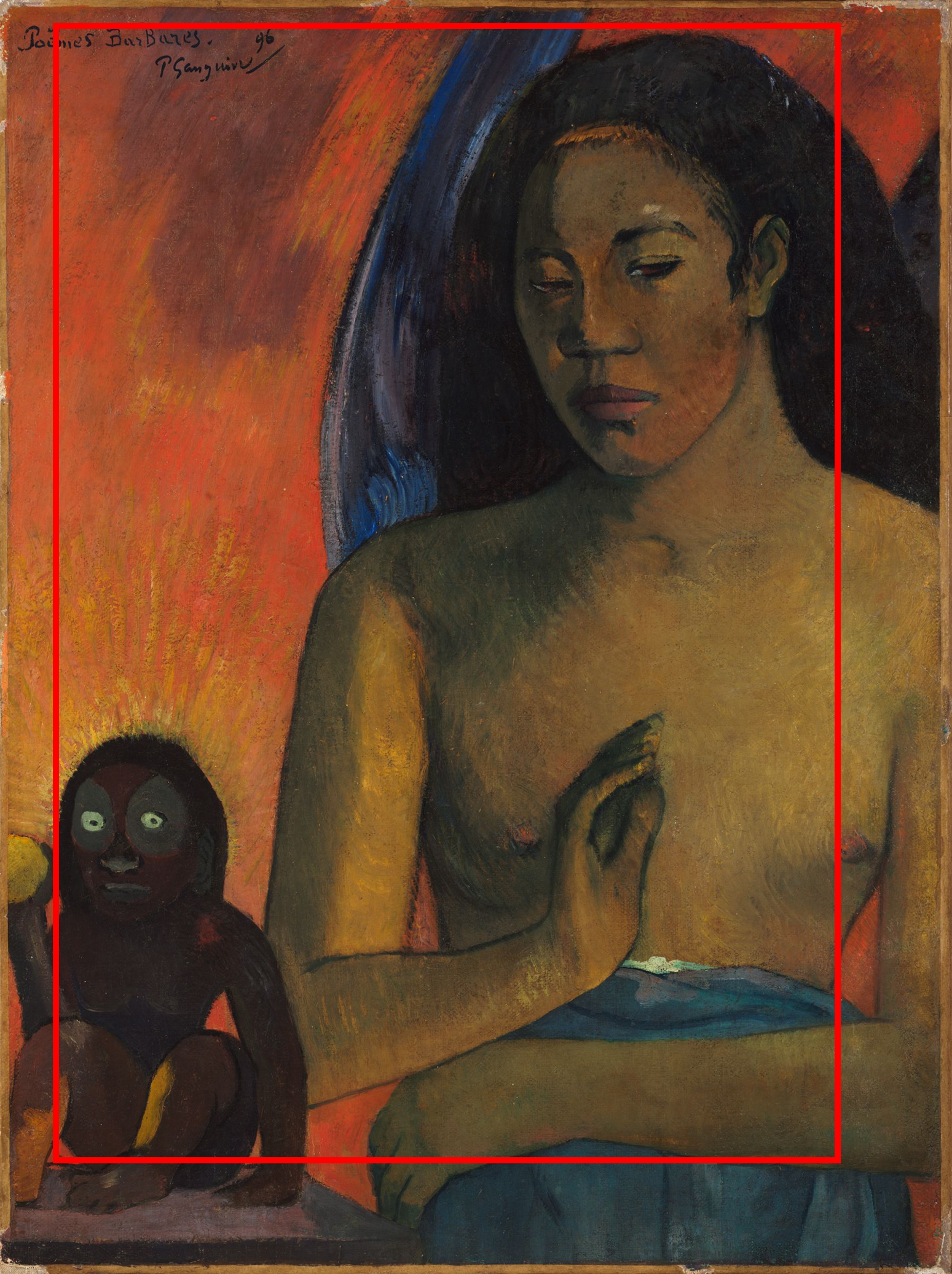}
 \caption{"Poèmes Barbares" (1896), oil on canvas, 64.8 x 48.3 cm (unframed), painted by the French artist Paul Gauguin (1848–1903), Harvard Art Museums/Fogg Museum, Bequest from the Collection of Maurice Wertheim, Class of 1906. Object Number: 1951.49 \textcopyright President and Fellows of Harvard College.}
 \label{fig:painting}
\end{figure}

Building on the ability to identify the pigments in the mock-up samples, we applied the models to the XRF dataset obtained from \textit{Poèmes Barbares} (Fig. \ref{fig:painting}), which was collected also by the XGLab ELIO XRF imaging spectrometer system~\cite{Vermeulen2021}. The red rectangle Fig. \ref{fig:painting} marks the area of investigation.

\begin{figure*}[htp!]
 \centering
 \includegraphics[width=\textwidth]{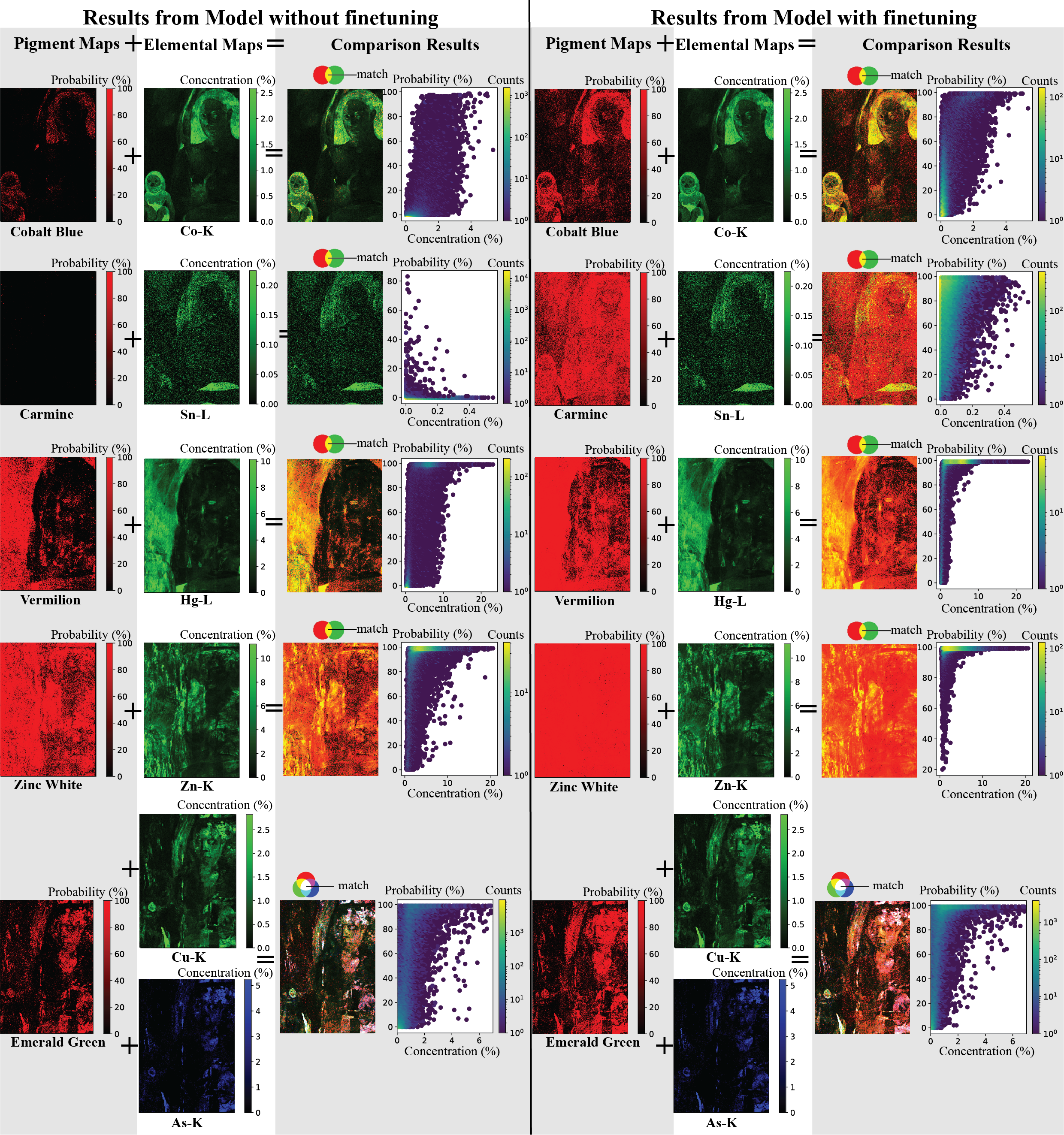}
 \caption{Pigment identification results of \textit{Poèmes Barbares} from the model without finetuning (left half), and the results from the finetuned model (right half), with results from one pigment on each row. The first column of images in each row shows the pigment map, with highest probability in red. The second column of images show the elemental map(s) for the pigment calculated by PyMCA, with highest concentrations in green or blue.  The third column images overlay the first two for comparison: it combines the red pigment map with the elements maps in green and (sometimes) blue. Yellow or white areas depict strong agreement between pigment maps and their corresponding elemental maps. The fourth column scatter plot compares  pigment probability and element concentration data for all image points, where element concentration sets the x-axis value (or for the bottom row, the minimum concentration of two elements), and the pigment probability sets the y-axis value. The right half of the figure shows how finetuning our DL model improves its results, and depicts images in the same arrangement used in the figure's left half. 
}
 \label{fig:GauguinResult}
\end{figure*}

As shown in Fig. \ref{fig:GauguinResult}, the pigment maps suggest the probability of pigments' existence as predicted by the pigment identification model, where an increased pixel brightness suggests a higher probability. The probability ranges from 0 - 100\%, calculated directly from  the XRF spectrum image. Their corresponding 2D elemental maps generated from PyMCA are also shown in Fig. \ref{fig:GauguinResult} for comparison, where a brighter pixel suggests a higher elemental concentration. In addition, each pigment map and its corresponding elemental map were merged to better visualize their pixel differences. Note here that as the elemental maps have a higher dynamic range than the pigment maps, we adjusted the max limit of the color bar to 99th percentile of the concentration data to balance the brightness between the two maps. To further evaluate the performance of the pigment identification model, Fig. \ref{fig:GauguinResult} also includes a series of scatter plots, which summarize the relationship between the predicted pigment probability and its actual elemental concentration at each pixel location. With a higher accuracy in identifying the pigments, the points remain higher in the plot. If the model identifies a pigment accurately, the probability (y value) will approach to zero at zero concentration, but will rapidly increase with any non-zero elemental concentration (x value). 

As shown in Fig.\ref{fig:GauguinResult}, the DL model trained and finetuned  solely on the mock-ups is applicable to the paintings and simulations as well. Fig. \ref{fig:GauguinResult} displays the results of six pigments: cobalt blue, carmine, vermilion, zinc white, and emerald green, each with a unique elemental spectral profile. By comparing the models with and without finetuning, the scatter plots suggest significant effect of finetuning on improving the sensitivity and accuracy in identifying all pigments, particularly for the situation of low elemental concentrations such as carmine. As for the remaining six pigments (chrome oxide green, chrome yellow, iron oxide, Prussian blue, red lead, lead white), the current model has not yet been able to distinguish the pigments that share the same elements (Fig. S2 and Fig. S3 in Supplementary Information), as discussed below.

\subsection{Tests on Paul Cezanne's \textit{The Bathers}}

\begin{figure}[htp!]
 \centering
 \includegraphics[height=4cm]{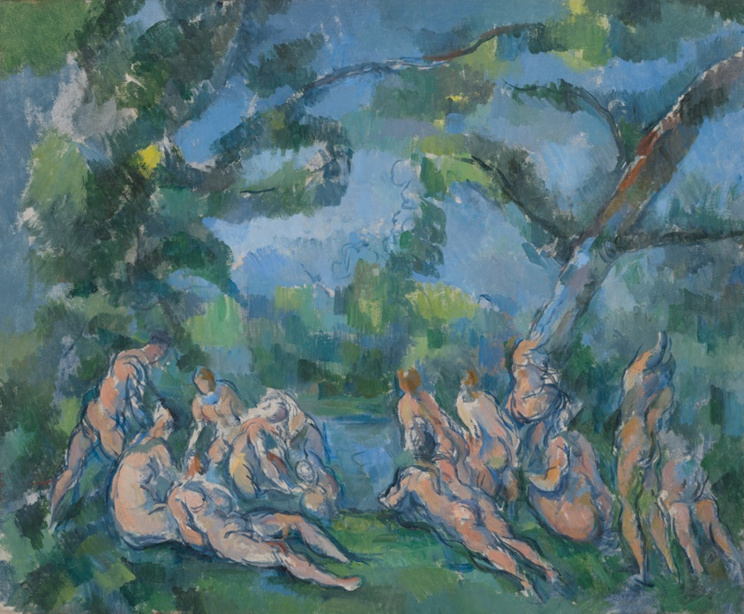}
 \caption{\textit{The Bathers} (1899-1904), oil on canvas, 51.3 × 61.7 cm, painted by the French artist Paul Cezanne (1839–1906), The Art Institute of Chicago, Amy McCormick Memorial Collection. Object Number: 1942.457 \textcopyright The Art Institute of Chicago.}
 \label{fig:painting2}
\end{figure}

\begin{figure*}[htp!] 
\begin{center}
\begin{tabular}{l} 
\includegraphics[width=\textwidth]{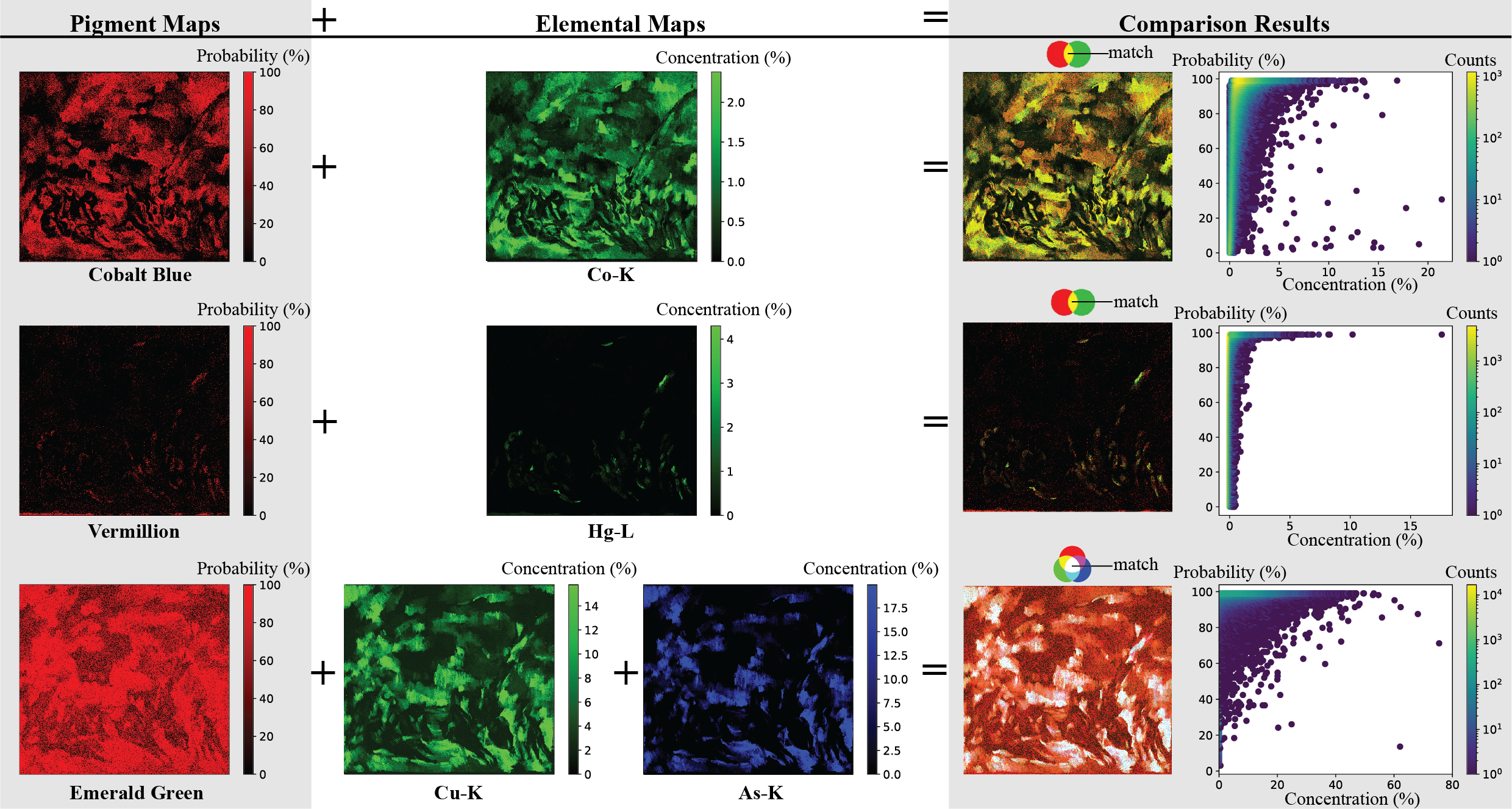}
\end{tabular}
\end{center}
\caption{Pigment identification results of \textit{The Bathers} from the finetuned model, including cobalt blue, vermilion, and emerald green from top to bottom. The first, second and third columns display the pigment map, element map(s) and the comparison figure and scatter plots, respectively.}
{\label{fig:CezanneResult} }
\end{figure*}

To further demonstrate that our model is also applicable to other paintings, we applied our same finetuned model to the XRF dataset of Paul Cezanne's \textit{The Bathers} (Fig. \ref{fig:painting2}). The XRF mapping was executed with a MA-XRF system at 40 kV and 1 A~\cite{vermeulen2022multi}, with an acquisition time of 100 ms per point and a step size of 1 mm. As shown in Fig. \ref{fig:CezanneResult}, pigment maps of cobalt blue, camine and emerald green perfectly match with their elemental maps and achieve high probabilities with concentrations increasing. The pigment identification of this painting shows results comparable to the Gauguin painting, highlighting the generalizability and stability of our model. A complete set of the elemental maps and the pigment maps with and without finetuning are given in Fig. S4-S6 in Supplementary Information. 

\FloatBarrier

\section{Discussion}
\subsection{Finetune the model}
Finetuning is a general technique popularized in deep learning models, especially on 2D images, to take advantage of weights trained on a huge dataset for another similar but smaller dataset. This technique has shown success in many fields, such as image recognition~\cite{ng2015deep}, medical diagnosis~\cite{khan2019novel} and unsupervised learning ~\cite{bengio2012deep}. In applying this method, the model initially learned the spectral features from the large simulation dataset followed by finetuning on the limited experimental data. Table \ref{tbl:result} suggests that finetuning, even using a small subset of the dataset, can significantly improve the performance of pigment identification. In particular for the cases of low elemental concentrations, such as carmine, the accuracy, sensitivity, and F1 score increased from 0.653 to 0.995, 0.664 to 0.992, and 0.685 to 0.995, respectively.

\begin{figure}[htp!]
 \centering
 \includegraphics[height=5cm]{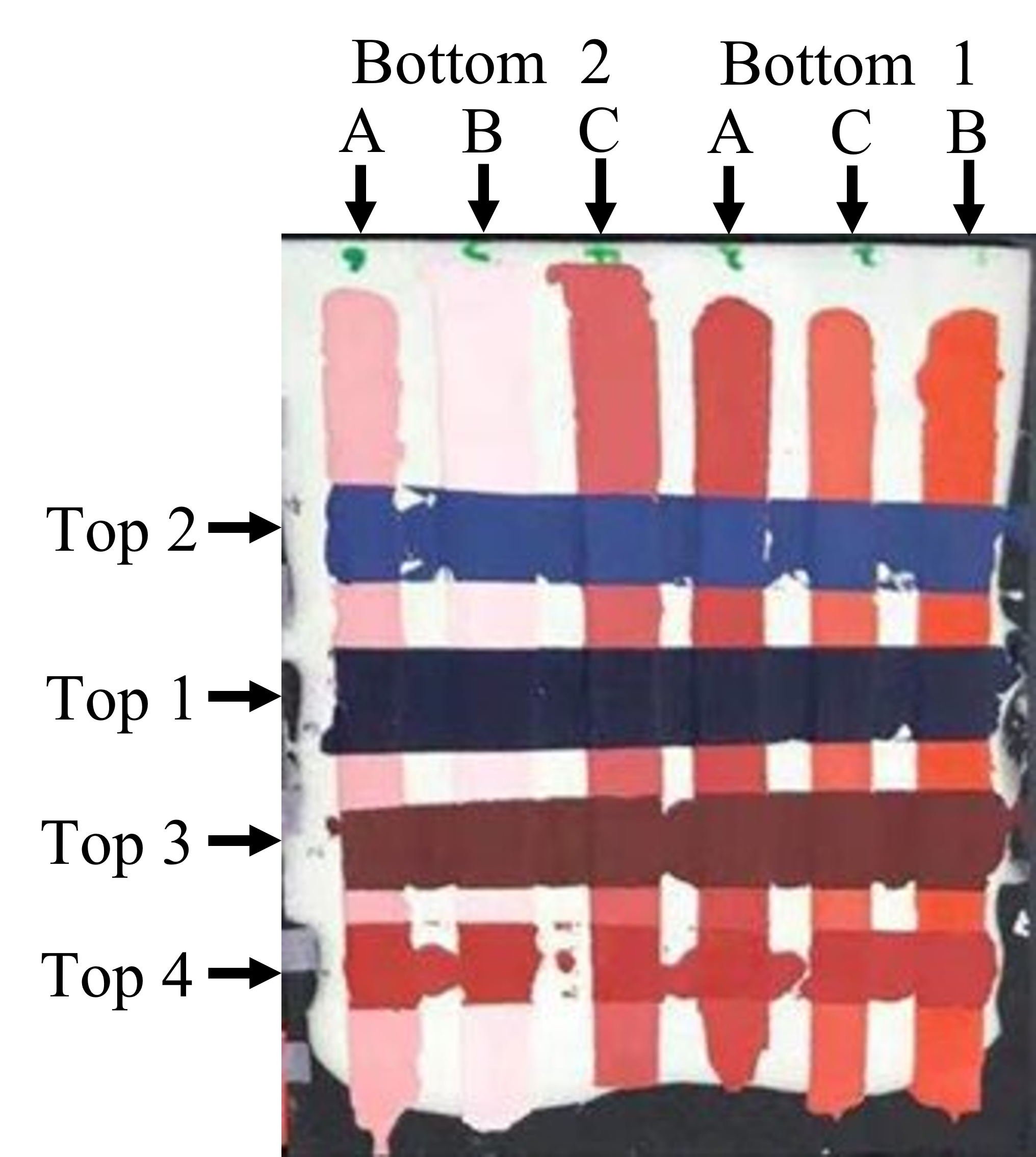}
 \caption{Photo image of one mock-up that contains Top 2, Top 1, Top 3, Top 4 (from top to bottom) as top layers and Bottom 2 with three different mass fractions (A, B, C), Bottom 1 with three different mass fractions (A, C, B) (from left to right) as bottom layers. }
 \label{fig:mockup}
\end{figure}

\begin{figure}[htp!]
 \centering

 \includegraphics[width=9cm]{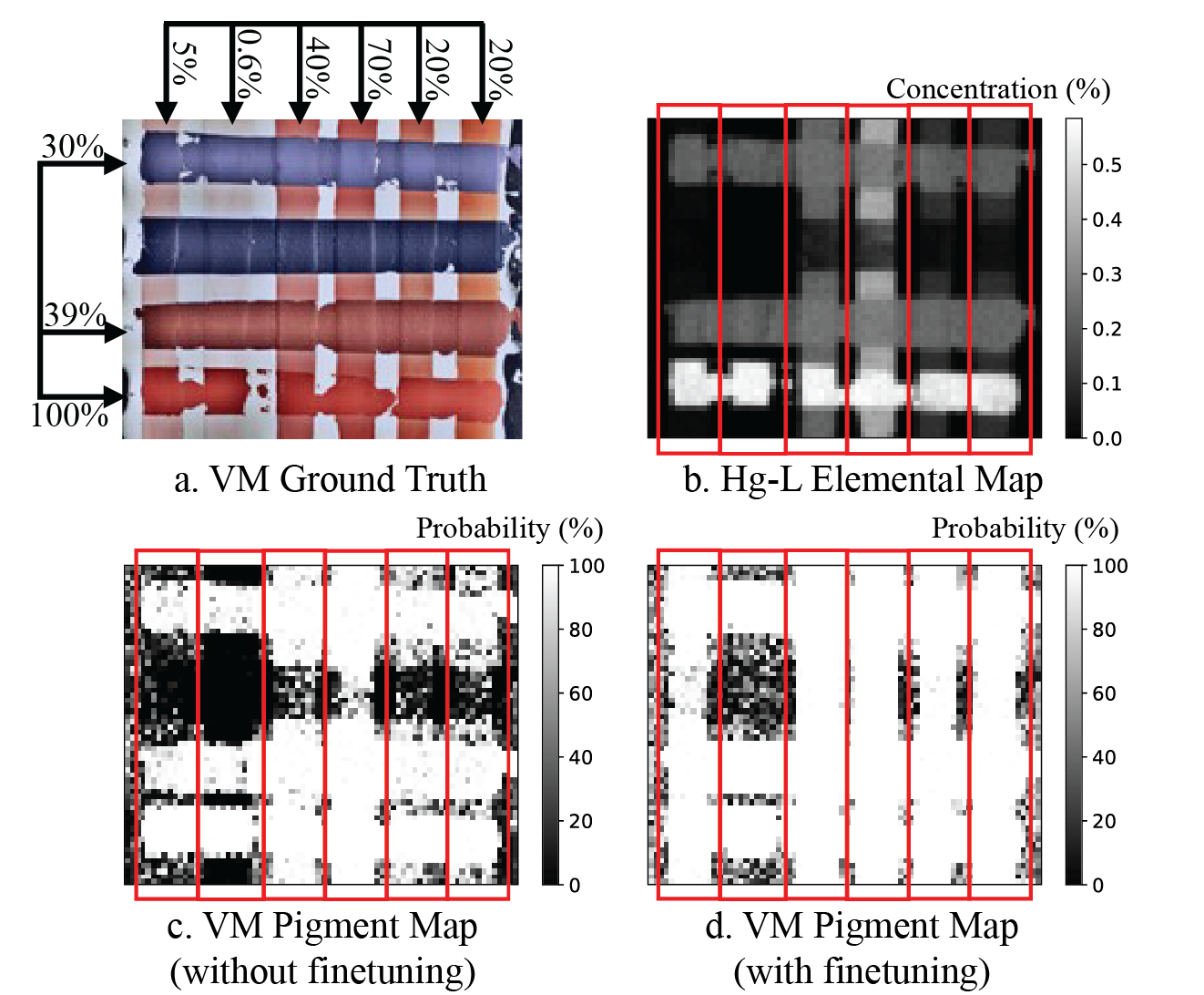}

 \caption{Comparison between pigment maps and elemental maps for the mock-up. (a) The ground truth of the location of vermilion (VM). (b) The Hg-L element map directly generated from PyMCA. (c) The VM pigment map generated from the model without finetuning identifies Hg in the 1, 3, and 4 rows and the 3 and 4 columns. It can barely detect Hg in the 5 and 6 columns. (d) The VM pigment map generated from the finetuned model indicates VM presence in the 1, 3, and 4 rows and 1, 3, 4, 5, and 6 columns with high probabilities. The only missing column of the VM pigment map is the second column, which contains 0.6\% of vermilion.}
 \label{fig:pigmentMap}
\end{figure}

To better visualize the effect of finetuning, we generated 2D pigment maps for one mock-up painting  (Fig. \ref{fig:mockup}) as an example. As shown in Fig. \ref{fig:pigmentMap}(a), the vermilion pigment (VM) was present in three horizontal paint strips (top pigment layer) and all six vertical strips (bottom pigment layer), but at different concentrations. However, neither the Hg elemental map nor the VM pigment map detected it reliably without finetuning. While Fig. \ref{fig:pigmentMap}(b) and (c)) may reveal the existence of VM at low concentrations, finetuning significantly improved the identification result of VM at low concentrations in Fig. \ref{fig:pigmentMap}(d), reaching its limit near 0.6\% concentration as shown in the second vertical strip.

To our best knowledge, this work is the first to apply transfer learning to pigment classification using XRF spectra. By finetuning on only $20\%$ randomly selected of the experimental data, we observed significant improvements in pigment identification. On one hand, it improved the model performance even with a limited training dataset. On the other hand, since mock-ups are hard to make, it releases the pressure of preparing a huge experimental dataset. However, as mentioned previously, since the pigment combination in the experimental dataset is limited, the finetuned model may overfit. Nevertheless, the finetuning technique contributes to extracting the features related to some specific pigment mixtures in the painter's palette, which helps the identification of pigments, especially in a specific painting, with similar painting styles. Therefore, this finetuning technique can be applied to many different fields using XRF spectra.
   
\subsection{Pigments with similar elemental profiles}

Three groups of pigments in our pigment library posed challenges due to similar elemental profiles: the chromium-containing group (chrome oxide green and chrome yellow), the iron-containing group (Prussian blue and iron oxide), and the lead-containing group (lead white and red lead). Although the two pigments in the chromium-containing group slightly vary in their elemental map (chrome oxide green only contains Cr and chrome yellow contains Cr and Pb), the model failed to distinguish between these two pigments. This is possibly caused by Pb, which exists almost everywhere in the painting, often mixed with other pigments (e.g. lead white). 

\begin{figure}[htp!]
 \centering

 \includegraphics[width=9cm]{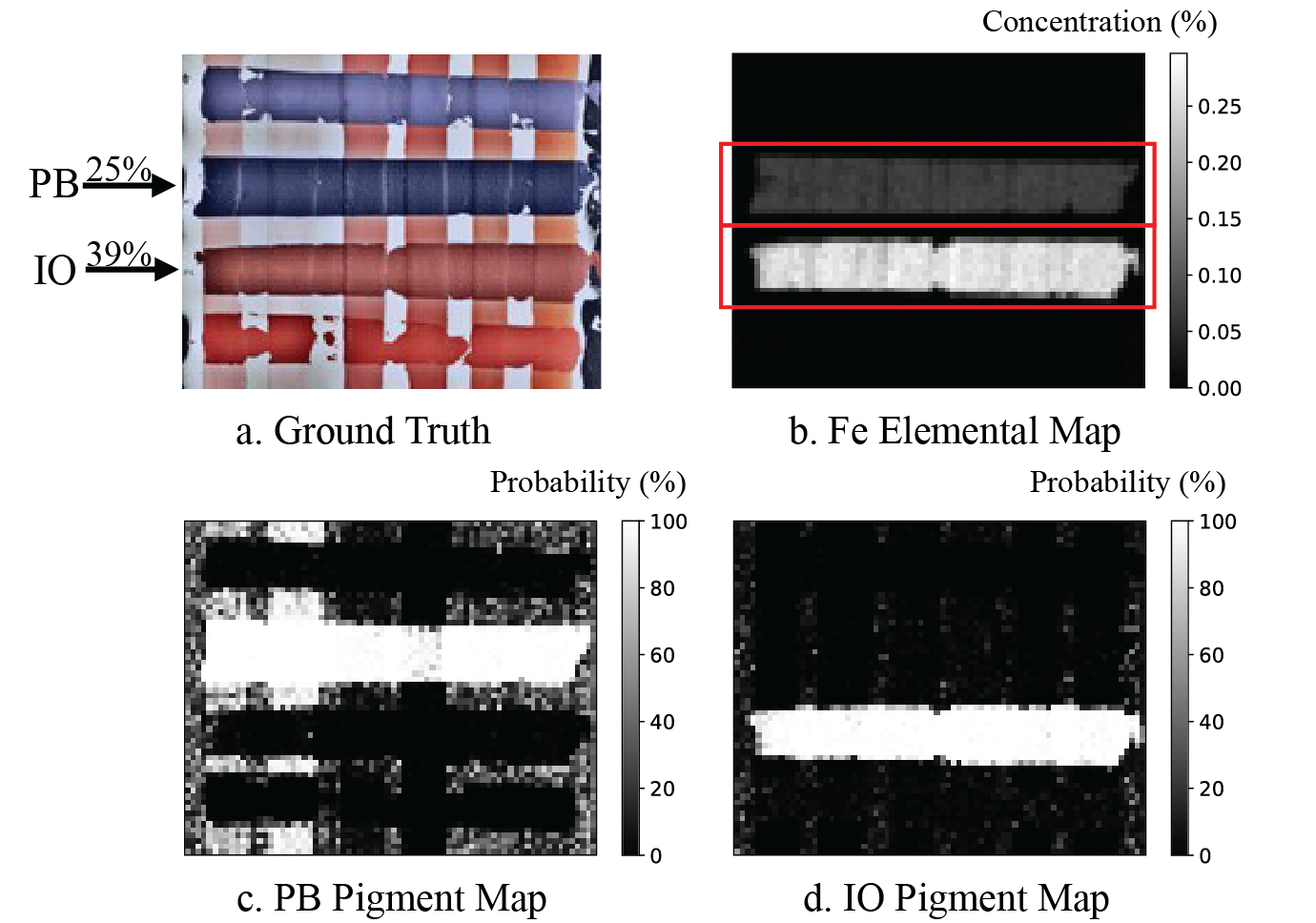}

 \caption{Our model distinguishes Prussian blue (PB) and iron oxide (IO) in one mock-up. (a) The ground truth of PB and IO locations. (b) The Fe element map directly  generated from PyMCA contains both PB and IO. (c) The PB pigment map generated from our model. (d) The IO pigment map generated from our model.}
 \label{fig:pigmentMap_Fe}
\end{figure}

For the other two pigment groups that share similar elemental profiles, the model before finetuning cannot distinguish them and shows low to medium probability for all areas that contain the element(s). However, the finetuned model can distinguish between those pigments when present in different pigment mixtures, but this only applies when the testing dataset contains the same pigment mixtures as the finetuning dataset. One special case in our result is that the finetune model can distinguish Prussian blue (PB) and iron oxide (IO) in our mock-ups. For example, in the mock-up in Fig. \ref{fig:mockup}, PB exists in the second row, and IO appears in the third row. (Fig. \ref{fig:pigmentMap_Fe}(a)). The only element detectable by XRF in Prussian blue and iron oxide is Fe. Therefore, the Fe element map (Fig. \ref{fig:pigmentMap_Fe}(b)) shows the location of both pigments and cannot be used to separate one from the other. However, Prussian blue pigment map (Fig. \ref{fig:pigmentMap_Fe}(c)) and iron oxide pigment map (Fig. \ref{fig:pigmentMap_Fe}(e)), generated from our fintuned model, show promising results of distinguishing these two pigments. The two pigment maps both have high probabilities at the ground truth location of the pigment and fairly low probabilities at the ground truth location of the other pigment. At the same time, we notice the pigment maps incorrectly identify where the Fe element has low concentration or does not exist, especially for the Prussian blue pigment map, which may be related to the high tinting strength, and therefore low concentration where used, of Prussian blue~\cite{Vermeulen2021, vermeulen2022multi, glinsman2004application}. This situation also appears in the Gauguin and Cezanne painting results: when the pigment concentration is low, the probability might vary from 0 to 100\%. We infer that this is due to the normalization process during the spectrum preprocessing. Specifically, the small peaks (low intensities) are enlarged after the normalization, enhancing both useful information and noise level. In the Fe element map (Fig. \ref{fig:pigmentMap_Fe}(b)), Fe concentration is much higher in iron oxide than in Prussian blue. Therefore, the iron oxide pigment map only has a few noisy points at the blank area (only the ground layer), while the Prussian blue pigment map shows high probabilities at the 1, 2, 5, 6 columns and the blank area.

\subsection{X-Ray absorbent materials}
\begin{figure}[htp!]
 \centering

 \includegraphics[height=6cm]{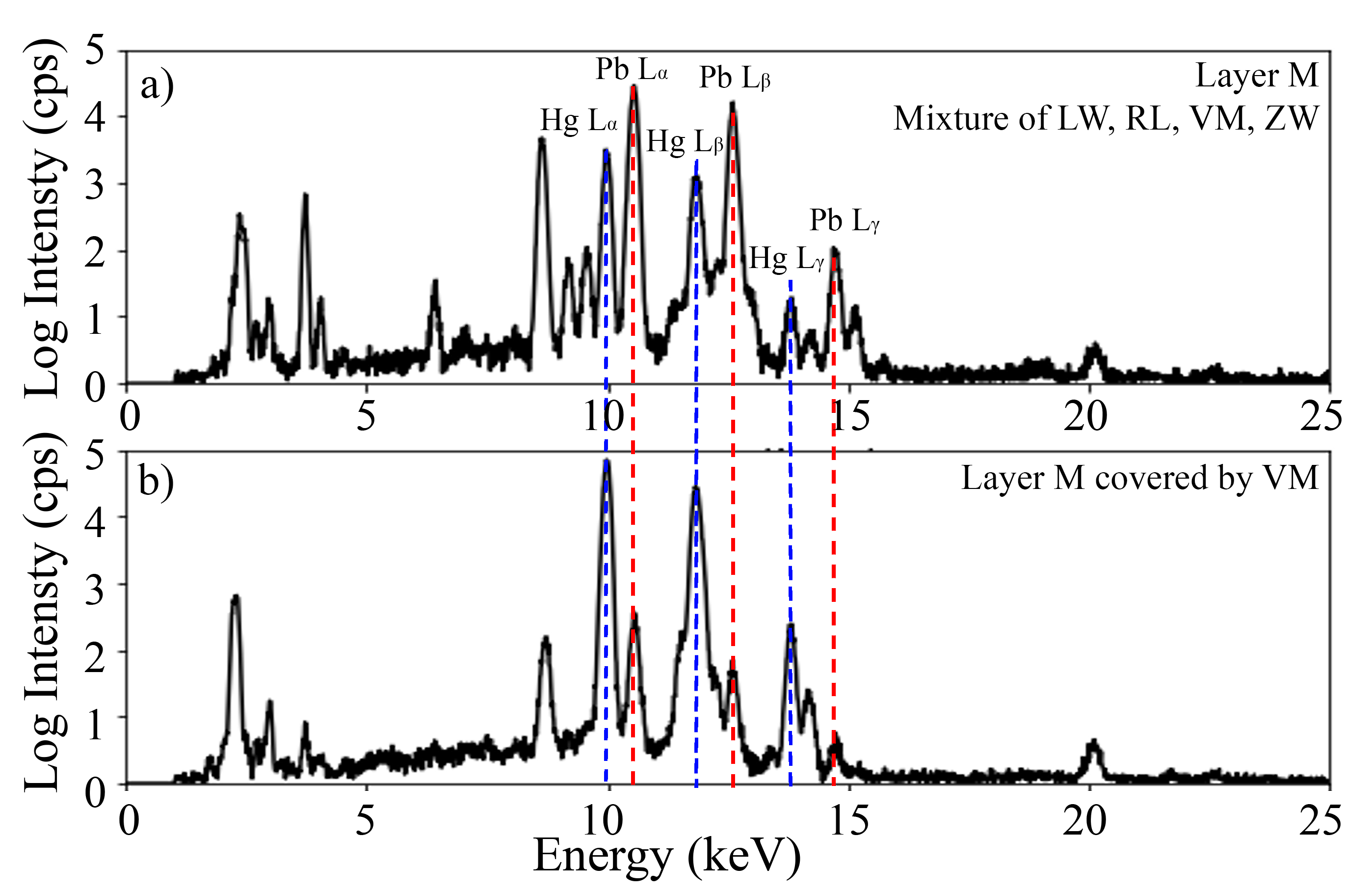}

 \caption{XRF spectra of the mock-up painting comparing the effect of highly absorbing pigments. a) One single layer M that contains lead white (LW), red lead (RL), vermilion (VM) and zinc white (ZW). The lead peaks marked in red have high intensities. b) The single layer M covered by one layer of vermilion (VM). The existence of VM in the top layer significantly blocks the XRF signal of the lead-containing pigments at the bottom, challenging our model's ability to detect the hidden lead element.}
 \label{fig:LowIntensity}
\end{figure}

Some pigments can strongly absorb photons at energies needed for X-ray fluorescence measurements, and absorption by top layers of a painting can severely distort the spectra detected from fluorescence of elements in the bottom layers. For example, the presence of vermilion in the top layer diminishes the lead peaks of lead white and red lead in the bottom layer (Fig. \ref{fig:LowIntensity}). In particular, as the energy of the Pb-L$\beta$ radiation is above the Hg-L3 edge, it is absorbed by vermilion in the top layer, decreasing the Pb-L$\beta$ to Pb-L$\alpha$ ratio. At the same time, the absorption gives rise to florescent emission from Hg-L$\alpha$, increasing the Hg-L$\alpha$ to Hg-L$\beta$ ratio. The shifts in line ratios for both elements create an extraordinary case the simulated data did not adequately describe and our model cannot distinguish from the underlying lead-containing pigments. This effect was previously observed in manual data evaluation as well~\cite{neelmeijer2000paintings}. 

\section{Conclusions}
XRF-based pigment identification problems have long required expert analysis and previous knowledge. In this paper, we pursued an automatic XRF data evaluation framework using deep learning. Our initial attempts at automatically identifying individual and overlapped pigments directly from XRF spectra show promise. While our model only tests a small number of pigments (11) and a small number of layers (2 + base) in this current stage, it automatically identified pigments in two different 19th-century paintings and in the training mock-ups and simulations they inspired. We intend this paper to stimulate further work in deep-learning assisted XRF studies for layered-pigment identification, and prompt more discussion of their feasibility and practicality for broader uses.

We focused on a set of representative pigments identified or considered present in Paul Gauguin's \textit{Poèmes Barbares} (1896)~\cite{Vermeulen2021} as a starting point and then created mock-ups to generate experimental datasets to capture nonlinear effects of layer structures. We added a simulation dataset to reduce the need for prohibitively tedious and difficult mock-up preparation. After data preprocessing, we trained the convolutional neural network with the simulation dataset and then finetuned it with the experimental dataset, therefore obtaining the pigment identification model.

Pigment maps are the visualization of the probability output of our model. The comparison of pigment maps and their corresponding element maps shows that our model can successfully identify pigments, especially in low concentration or in overpainted layers. However, the model still has some shortcomings: (a) The model cannot always distinguish pigments with similar elemental profiles; (b) It does not work when high absorbing pigments block the radiation emitted from the hidden layer; (c) The finetuned model needs experimental data with at least a small set of ground truth measurements to prevent limited pigment mixtures in the mock-ups from causing model overfitting and wrong predictions.

This research is still in its early stages, and there are multiple directions to extend the current work. First of all, including data from other techniques, such as spectral imaging to get molecular structures, might better differentiate pigments and compensate for shortcomings (a) and (b) which are common problems with using XRF to identify pigments. Second, we only try to identify presence or absence of pigments, and not their depth or the layers' sequence from front-to-back, an important but much more challenging problem. Third, the current pigment library is limited, and training the model with a larger range of pigments can make it more accessible to different paintings. Last but not least, some unsupervised learning methods have been proposed to extract latent features of signals without any labels -- an interesting approach to try in future work.

\section*{Author Contributions}

Conceptualization, Marc Walton, Aggelos Katsaggelos, Jack Tumblin, Florian Willomitzer, Matthias Alfeld and Bingjie (Jenny) Xu; Data curation, Pengxiao Hao, Marc Vermeulen and Alicia McGeachy; Formal analysis, Bingjie (Jenny) Xu, Pengxiao Hao, Yunan Wu and Marc Vermeulen; Funding acquisition, Marc Walton; Investigation, Bingjie (Jenny) Xu, Pengxiao Hao, Marc Vermeulen and Alicia McGeachy; Methodology, Bingjie (Jenny) Xu, Yunan Wu, Pengxiao Hao, Marc Vermeulen, Matthias Alfeld, Aggelos Katsaggelos and Marc Walton; Resources, Kate Smith, Katherine Eremin, Georgina Rayner, Giovanni Verri; Software: Bingjie Xu, Yunan Wu, Marc Vermeulen, Alicia McGeachy and Marc Walton; Writing – original draft, Bingjie (Jenny) Xu, Yunan Wu, Pengxiao Hao and Matthias Alfeld; Writing – review \& editing, Marc Vermeulen, Alicia McGeachy, Florian Willomitzer, Matthias Alfeld, Jack Tumblin,  Aggelos Katsaggelos and Marc Walton. All authors have read and agreed to the published version of the manuscript.



\section*{Acknowledgements}
The analysis of the Gauguin and Cezanne paintings is part of NU-ACCESS’s broad portfolio of activities, made possible by generous support of the Andrew W. Mellon Foundation as well as supplemental support provided by the Materials Research Center, the Office of the Vice President for Research, the McCormick School of Engineering and Applied Science and the Department of Materials Science and Engineering at Northwestern University. The authors gratefully acknowledge Emeline Pouyet and Gianluca Pastorelli (formerly NU-ACCESS) for the acquisition of the MA-XRF data on Gauguin’s Poèmes Barbares.



\balance


\bibliography{reference} 

\providecommand*{\mcitethebibliography}{\thebibliography}
\csname @ifundefined\endcsname{endmcitethebibliography}
{\let\endmcitethebibliography\endthebibliography}{}
\begin{mcitethebibliography}{34}
\providecommand*{\natexlab}[1]{#1}
\providecommand*{\mciteSetBstSublistMode}[1]{}
\providecommand*{\mciteSetBstMaxWidthForm}[2]{}
\providecommand*{\mciteBstWouldAddEndPuncttrue}
  {\def\EndOfBibitem{\unskip.}}
\providecommand*{\mciteBstWouldAddEndPunctfalse}
  {\let\EndOfBibitem\relax}
\providecommand*{\mciteSetBstMidEndSepPunct}[3]{}
\providecommand*{\mciteSetBstSublistLabelBeginEnd}[3]{}
\providecommand*{\EndOfBibitem}{}
\mciteSetBstSublistMode{f}
\mciteSetBstMaxWidthForm{subitem}
{(\emph{\alph{mcitesubitemcount}})}
\mciteSetBstSublistLabelBeginEnd{\mcitemaxwidthsubitemform\space}
{\relax}{\relax}

\bibitem[van Hoof \emph{et~al.}(2021)van Hoof, Bacon, Fittschen, and
  Vincze]{Vanhoof2021}
C.~van Hoof, J.~R. Bacon, U.~E.~A. Fittschen and L.~Vincze, \emph{Journal of
  Analytical Atomic Spectrometry}, 2021, \textbf{36}, 1797--1812\relax
\mciteBstWouldAddEndPuncttrue
\mciteSetBstMidEndSepPunct{\mcitedefaultmidpunct}
{\mcitedefaultendpunct}{\mcitedefaultseppunct}\relax
\EndOfBibitem
\bibitem[Rowe \emph{et~al.}(2012)Rowe, Hughes, and
  Robinson]{rowe2012quantification}
H.~Rowe, N.~Hughes and K.~Robinson, \emph{Chemical Geology}, 2012,
  \textbf{324}, 122--131\relax
\mciteBstWouldAddEndPuncttrue
\mciteSetBstMidEndSepPunct{\mcitedefaultmidpunct}
{\mcitedefaultendpunct}{\mcitedefaultseppunct}\relax
\EndOfBibitem
\bibitem[Oyedotun(2018)]{oyedotun2018x}
T.~D.~T. Oyedotun, \emph{Geology, Ecology, and Landscapes}, 2018, \textbf{2},
  148--154\relax
\mciteBstWouldAddEndPuncttrue
\mciteSetBstMidEndSepPunct{\mcitedefaultmidpunct}
{\mcitedefaultendpunct}{\mcitedefaultseppunct}\relax
\EndOfBibitem
\bibitem[Sarala(2016)]{sarala2016comparison}
P.~Sarala, \emph{Geochemistry: Exploration, Environment, Analysis}, 2016,
  \textbf{16}, 181--192\relax
\mciteBstWouldAddEndPuncttrue
\mciteSetBstMidEndSepPunct{\mcitedefaultmidpunct}
{\mcitedefaultendpunct}{\mcitedefaultseppunct}\relax
\EndOfBibitem
\bibitem[Langstraat \emph{et~al.}(2017)Langstraat, Knijnenberg, Edelman, Van
  De~Merwe, van Loon, Dik, and van Asten]{langstraat2017large}
K.~Langstraat, A.~Knijnenberg, G.~Edelman, L.~Van De~Merwe, A.~van Loon, J.~Dik
  and A.~van Asten, \emph{Scientific reports}, 2017, \textbf{7}, 1--11\relax
\mciteBstWouldAddEndPuncttrue
\mciteSetBstMidEndSepPunct{\mcitedefaultmidpunct}
{\mcitedefaultendpunct}{\mcitedefaultseppunct}\relax
\EndOfBibitem
\bibitem[Nakano \emph{et~al.}(2011)Nakano, Nishi, Otsuki, Nishiwaki, and
  Tsuji]{nakano2011depth}
K.~Nakano, C.~Nishi, K.~Otsuki, Y.~Nishiwaki and K.~Tsuji, \emph{Analytical
  chemistry}, 2011, \textbf{83}, 3477--3483\relax
\mciteBstWouldAddEndPuncttrue
\mciteSetBstMidEndSepPunct{\mcitedefaultmidpunct}
{\mcitedefaultendpunct}{\mcitedefaultseppunct}\relax
\EndOfBibitem
\bibitem[Shackley(2011)]{shackley2011introduction}
M.~S. Shackley, \emph{X-Ray Fluorescence spectrometry (XRF) in geoarchaeology},
  Springer, 2011, pp. 7--44\relax
\mciteBstWouldAddEndPuncttrue
\mciteSetBstMidEndSepPunct{\mcitedefaultmidpunct}
{\mcitedefaultendpunct}{\mcitedefaultseppunct}\relax
\EndOfBibitem
\bibitem[Alfeld and de~Viguerie(2017)]{Alfeld2017}
M.~Alfeld and L.~de~Viguerie, \emph{Spectrochimica Acta Part B: Atomic
  Spectroscopy}, 2017, \textbf{136}, 81--105\relax
\mciteBstWouldAddEndPuncttrue
\mciteSetBstMidEndSepPunct{\mcitedefaultmidpunct}
{\mcitedefaultendpunct}{\mcitedefaultseppunct}\relax
\EndOfBibitem
\bibitem[van Espen and Janssens(1993)]{van1993spectrum}
P.~J. van Espen and K.~H. Janssens, \emph{Handbook of X-Ray spectrometry:
  Methods and techniques}, Marcel Dekker, Inc., New York, NY, 1993, ch.~5, pp.
  181--293\relax
\mciteBstWouldAddEndPuncttrue
\mciteSetBstMidEndSepPunct{\mcitedefaultmidpunct}
{\mcitedefaultendpunct}{\mcitedefaultseppunct}\relax
\EndOfBibitem
\bibitem[Sol{\'e} \emph{et~al.}(2007)Sol{\'e}, Papillon, Cotte, Walter, and
  Susini]{sole2007multiplatform}
V.~Sol{\'e}, E.~Papillon, M.~Cotte, P.~Walter and J.~Susini,
  \emph{Spectrochimica Acta Part B: Atomic Spectroscopy}, 2007, \textbf{62},
  63--68\relax
\mciteBstWouldAddEndPuncttrue
\mciteSetBstMidEndSepPunct{\mcitedefaultmidpunct}
{\mcitedefaultendpunct}{\mcitedefaultseppunct}\relax
\EndOfBibitem
\bibitem[Alfeld and Janssens(2015)]{alfeld2015strategies}
M.~Alfeld and K.~Janssens, \emph{Journal of analytical atomic spectrometry},
  2015, \textbf{30}, 777--789\relax
\mciteBstWouldAddEndPuncttrue
\mciteSetBstMidEndSepPunct{\mcitedefaultmidpunct}
{\mcitedefaultendpunct}{\mcitedefaultseppunct}\relax
\EndOfBibitem
\bibitem[Romano \emph{et~al.}(2017)Romano, Caliri, Nicotra, Di~Martino,
  Pappalardo, Rizzo, and Santos]{romano2017real}
F.~P. Romano, C.~Caliri, P.~Nicotra, S.~Di~Martino, L.~Pappalardo, F.~Rizzo and
  H.~C. Santos, \emph{Journal of Analytical Atomic Spectrometry}, 2017,
  \textbf{32}, 773--781\relax
\mciteBstWouldAddEndPuncttrue
\mciteSetBstMidEndSepPunct{\mcitedefaultmidpunct}
{\mcitedefaultendpunct}{\mcitedefaultseppunct}\relax
\EndOfBibitem
\bibitem[Alfeld \emph{et~al.}(2013)Alfeld, De~Nolf, Cagno, Appel, Siddons,
  Kuczewski, Janssens, Dik, Trentelman,
  Walton,\emph{et~al.}]{alfeld2013revealing}
M.~Alfeld, W.~De~Nolf, S.~Cagno, K.~Appel, D.~P. Siddons, A.~Kuczewski,
  K.~Janssens, J.~Dik, K.~Trentelman, M.~Walton \emph{et~al.}, \emph{Journal of
  Analytical Atomic Spectrometry}, 2013, \textbf{28}, 40--51\relax
\mciteBstWouldAddEndPuncttrue
\mciteSetBstMidEndSepPunct{\mcitedefaultmidpunct}
{\mcitedefaultendpunct}{\mcitedefaultseppunct}\relax
\EndOfBibitem
\bibitem[Kogou \emph{et~al.}(2021)Kogou, Lee, Shahtahmassebi, and
  Liang]{kogou2021new}
S.~Kogou, L.~Lee, G.~Shahtahmassebi and H.~Liang, \emph{X-Ray Spectrometry},
  2021, \textbf{50}, 310--319\relax
\mciteBstWouldAddEndPuncttrue
\mciteSetBstMidEndSepPunct{\mcitedefaultmidpunct}
{\mcitedefaultendpunct}{\mcitedefaultseppunct}\relax
\EndOfBibitem
\bibitem[Vermeulen \emph{et~al.}(2022)Vermeulen, McGeachy, Xu, Chopp,
  Kataggelos, Meyers, Alfeld, and Walton]{XRFast}
M.~Vermeulen, A.~McGeachy, B.~Xu, H.~Chopp, A.~Kataggelos, R.~Meyers, M.~Alfeld
  and M.~Walton, Manuscript submitted for publication.\relax
\mciteBstWouldAddEndPunctfalse
\mciteSetBstMidEndSepPunct{\mcitedefaultmidpunct}
{}{\mcitedefaultseppunct}\relax
\EndOfBibitem
\bibitem[Shugar \emph{et~al.}(2021)Shugar, Drake, and Kelley]{shugar2021rapid}
A.~N. Shugar, B.~L. Drake and G.~Kelley, \emph{Scientific reports}, 2021,
  \textbf{11}, 1--10\relax
\mciteBstWouldAddEndPuncttrue
\mciteSetBstMidEndSepPunct{\mcitedefaultmidpunct}
{\mcitedefaultendpunct}{\mcitedefaultseppunct}\relax
\EndOfBibitem
\bibitem[Kim \emph{et~al.}(2022)Kim, Ling, Plattenberger, Clarens, and
  Peters]{kim2022quantification}
J.~J. Kim, F.~T. Ling, D.~A. Plattenberger, A.~F. Clarens and C.~A. Peters,
  \emph{Applied Geochemistry}, 2022, \textbf{136}, 105162\relax
\mciteBstWouldAddEndPuncttrue
\mciteSetBstMidEndSepPunct{\mcitedefaultmidpunct}
{\mcitedefaultendpunct}{\mcitedefaultseppunct}\relax
\EndOfBibitem
\bibitem[Jones \emph{et~al.}(2022)Jones, Daly, Higgitt, and
  Rodrigues]{jones2022neural}
C.~Jones, N.~S. Daly, C.~Higgitt and M.~R. Rodrigues, \emph{Heritage Science},
  2022, \textbf{10}, 1--14\relax
\mciteBstWouldAddEndPuncttrue
\mciteSetBstMidEndSepPunct{\mcitedefaultmidpunct}
{\mcitedefaultendpunct}{\mcitedefaultseppunct}\relax
\EndOfBibitem
\bibitem[Vermeulen \emph{et~al.}(2021)Vermeulen, Smith, Eremin, Rayner, and
  Walton]{Vermeulen2021}
M.~Vermeulen, K.~Smith, K.~Eremin, G.~Rayner and M.~Walton,
  \emph{Spectrochimica Acta Part A: Molecular and Biomolecular Spectroscopy},
  2021, \textbf{252}, 119547\relax
\mciteBstWouldAddEndPuncttrue
\mciteSetBstMidEndSepPunct{\mcitedefaultmidpunct}
{\mcitedefaultendpunct}{\mcitedefaultseppunct}\relax
\EndOfBibitem
\bibitem[Sturdy(2016)]{Sturdy2016}
L.~F. Sturdy, \emph{PhD thesis}, Northwestern University, 2016\relax
\mciteBstWouldAddEndPuncttrue
\mciteSetBstMidEndSepPunct{\mcitedefaultmidpunct}
{\mcitedefaultendpunct}{\mcitedefaultseppunct}\relax
\EndOfBibitem
\bibitem[De~Viguerie \emph{et~al.}(2009)De~Viguerie, Sole, and
  Walter]{de2009multilayers}
L.~De~Viguerie, V.~A. Sole and P.~Walter, \emph{Analytical and bioanalytical
  chemistry}, 2009, \textbf{395}, 2015--2020\relax
\mciteBstWouldAddEndPuncttrue
\mciteSetBstMidEndSepPunct{\mcitedefaultmidpunct}
{\mcitedefaultendpunct}{\mcitedefaultseppunct}\relax
\EndOfBibitem
\bibitem[Ryan \emph{et~al.}(1988)Ryan, Clayton, Griffin, Sie, and
  Cousens]{RYAN1988396}
C.~Ryan, E.~Clayton, W.~Griffin, S.~Sie and D.~Cousens, \emph{Nuclear
  Instruments and Methods in Physics Research Section B: Beam Interactions with
  Materials and Atoms}, 1988, \textbf{34}, 396--402\relax
\mciteBstWouldAddEndPuncttrue
\mciteSetBstMidEndSepPunct{\mcitedefaultmidpunct}
{\mcitedefaultendpunct}{\mcitedefaultseppunct}\relax
\EndOfBibitem
\bibitem[Xu \emph{et~al.}(2015)Xu, Wang, Chen, and Li]{xu2015empirical}
B.~Xu, N.~Wang, T.~Chen and M.~Li, \emph{arXiv preprint arXiv:1505.00853},
  2015\relax
\mciteBstWouldAddEndPuncttrue
\mciteSetBstMidEndSepPunct{\mcitedefaultmidpunct}
{\mcitedefaultendpunct}{\mcitedefaultseppunct}\relax
\EndOfBibitem
\bibitem[Zhuang \emph{et~al.}(2020)Zhuang, Qi, Duan, Xi, Zhu, Zhu, Xiong, and
  He]{zhuang2020comprehensive}
F.~Zhuang, Z.~Qi, K.~Duan, D.~Xi, Y.~Zhu, H.~Zhu, H.~Xiong and Q.~He,
  \emph{Proceedings of the IEEE}, 2020, \textbf{109}, 43--76\relax
\mciteBstWouldAddEndPuncttrue
\mciteSetBstMidEndSepPunct{\mcitedefaultmidpunct}
{\mcitedefaultendpunct}{\mcitedefaultseppunct}\relax
\EndOfBibitem
\bibitem[Raghu \emph{et~al.}(2019)Raghu, Zhang, Kleinberg, and
  Bengio]{raghu2019transfusion}
M.~Raghu, C.~Zhang, J.~Kleinberg and S.~Bengio, Advances in Neural Information
  Processing Systems, 2019\relax
\mciteBstWouldAddEndPuncttrue
\mciteSetBstMidEndSepPunct{\mcitedefaultmidpunct}
{\mcitedefaultendpunct}{\mcitedefaultseppunct}\relax
\EndOfBibitem
\bibitem[Stone(1974)]{stone1974cross}
M.~Stone, \emph{Journal of the royal statistical society: Series B
  (Methodological)}, 1974, \textbf{36}, 111--133\relax
\mciteBstWouldAddEndPuncttrue
\mciteSetBstMidEndSepPunct{\mcitedefaultmidpunct}
{\mcitedefaultendpunct}{\mcitedefaultseppunct}\relax
\EndOfBibitem
\bibitem[Kingma and Ba(2014)]{kingma2014adam}
D.~P. Kingma and J.~Ba, \emph{arXiv preprint arXiv:1412.6980}, 2014\relax
\mciteBstWouldAddEndPuncttrue
\mciteSetBstMidEndSepPunct{\mcitedefaultmidpunct}
{\mcitedefaultendpunct}{\mcitedefaultseppunct}\relax
\EndOfBibitem
\bibitem[Dapson(2007)]{Dapson2007}
R.~Dapson, \emph{Biotechnic \& Histochemistry}, 2007, \textbf{82},
  173--187\relax
\mciteBstWouldAddEndPuncttrue
\mciteSetBstMidEndSepPunct{\mcitedefaultmidpunct}
{\mcitedefaultendpunct}{\mcitedefaultseppunct}\relax
\EndOfBibitem
\bibitem[Vermeulen \emph{et~al.}(2022)Vermeulen, Miranda, Tamburini, Delgado,
  and Walton]{vermeulen2022multi}
M.~Vermeulen, A.~S.~O. Miranda, D.~Tamburini, S.~E.~R. Delgado and M.~Walton,
  \emph{Heritage Science}, 2022, \textbf{10}, 1--22\relax
\mciteBstWouldAddEndPuncttrue
\mciteSetBstMidEndSepPunct{\mcitedefaultmidpunct}
{\mcitedefaultendpunct}{\mcitedefaultseppunct}\relax
\EndOfBibitem
\bibitem[Ng \emph{et~al.}(2015)Ng, Nguyen, Vonikakis, and Winkler]{ng2015deep}
H.-W. Ng, V.~D. Nguyen, V.~Vonikakis and S.~Winkler, Proceedings of the 2015
  ACM on international conference on multimodal interaction, 2015, pp.
  443--449\relax
\mciteBstWouldAddEndPuncttrue
\mciteSetBstMidEndSepPunct{\mcitedefaultmidpunct}
{\mcitedefaultendpunct}{\mcitedefaultseppunct}\relax
\EndOfBibitem
\bibitem[Khan \emph{et~al.}(2019)Khan, Islam, Jan, Din, and
  Rodrigues]{khan2019novel}
S.~Khan, N.~Islam, Z.~Jan, I.~U. Din and J.~J.~C. Rodrigues, \emph{Pattern
  Recognition Letters}, 2019, \textbf{125}, 1--6\relax
\mciteBstWouldAddEndPuncttrue
\mciteSetBstMidEndSepPunct{\mcitedefaultmidpunct}
{\mcitedefaultendpunct}{\mcitedefaultseppunct}\relax
\EndOfBibitem
\bibitem[Bengio(2012)]{bengio2012deep}
Y.~Bengio, Proceedings of ICML workshop on unsupervised and transfer learning,
  2012, pp. 17--36\relax
\mciteBstWouldAddEndPuncttrue
\mciteSetBstMidEndSepPunct{\mcitedefaultmidpunct}
{\mcitedefaultendpunct}{\mcitedefaultseppunct}\relax
\EndOfBibitem
\bibitem[Glinsman
  \emph{et~al.}(2004)Glinsman\emph{et~al.}]{glinsman2004application}
L.~D. Glinsman \emph{et~al.}, \emph{PhD thesis}, Universiteit van Amsterdam
  [Host], 2004\relax
\mciteBstWouldAddEndPuncttrue
\mciteSetBstMidEndSepPunct{\mcitedefaultmidpunct}
{\mcitedefaultendpunct}{\mcitedefaultseppunct}\relax
\EndOfBibitem
\bibitem[Neelmeijer \emph{et~al.}(2000)Neelmeijer, Brissaud, Calligaro,
  Demortier, Hautoj{\"a}rvi, M{\"a}der, Martinot, Schreiner, Tuurnala, and
  Weber]{neelmeijer2000paintings}
C.~Neelmeijer, I.~Brissaud, T.~Calligaro, G.~Demortier, A.~Hautoj{\"a}rvi,
  M.~M{\"a}der, L.~Martinot, M.~Schreiner, T.~Tuurnala and G.~Weber,
  \emph{X-Ray Spectrometry: An International Journal}, 2000, \textbf{29},
  101--110\relax
\mciteBstWouldAddEndPuncttrue
\mciteSetBstMidEndSepPunct{\mcitedefaultmidpunct}
{\mcitedefaultendpunct}{\mcitedefaultseppunct}\relax
\EndOfBibitem
\end{mcitethebibliography}
\bibliographystyle{rsc} 

\end{document}